\documentclass[runningheads]{llncs}
\usepackage{graphicx}
\usepackage{amsmath,amssymb} 
\usepackage{color}
\usepackage{xcolor} 
\usepackage{subfigure}
\usepackage{tikz}
\usepackage{hhline}
\usepackage{makecell}
\usepackage{multirow} 
\usepackage[width=122mm,left=12mm,paperwidth=146mm,height=193mm,top=12mm,paperheight=217mm]{geometry}
\usepackage{ulem}

\begin{document}
\newcommand{\tce}[1]{\textcolor{orange}{\textbf{#1}}}	      
\newcommand{\tcb}[1]{\textcolor{green}{\textbf{#1}}}	  
\newcommand{\tct}[1]{\textcolor{brown}{\textbf{#1}}}  
\newcommand{\tcz}[1]{\textcolor{magenta}{\textbf{#1}}}  
\newcommand{\todo}{$\clubsuit$} 
\newcommand{\TODO}{$\clubsuit$\textbf{TODO}$\clubsuit$}

\renewcommand\theadalign{bc}
\renewcommand\theadfont{\bfseries}
\renewcommand\theadgape{\Gape[4pt]}
\renewcommand\cellgape{\Gape[4pt]}

\newcommand{\FS}{\mathrm{S}}
\newcommand{\FC}{\mathrm{C}}

\newcommand{\etal}{{et al.}}
\newcommand{\etall}{{et al.}} 

\newcommand{\pz}{\phantom{0}}%

\newcommand{\labelimage}[3]{
    \begin{tikzpicture}
        \node[inner sep=0, anchor=south west] (image) at (0,0) {\includegraphics[width=#1\textwidth]{#2}};
        \node[inner sep=0, anchor=north east,white,fill=black,align=right] at (image.north east) {\tiny\textsf{#3}};
    \end{tikzpicture}
}


\title{
Occlusions, Motion and Depth Boundaries with a Generic Network 
for Disparity, Optical Flow or Scene Flow Estimation 
}

\titlerunning{Occlusions, Motion and Depth Boundaries}

\author{Eddy Ilg\textsuperscript{*} \and 
Tonmoy Saikia\textsuperscript{*} \and
Margret Keuper \and 
Thomas Brox}
\institute{
University of Freiburg, Germany \\
\email{\{ilg,saikia,keuper,brox\}@cs.uni-freiburg.de}
}

\authorrunning{E. Ilg, T. Saikia, M. Keuper and T. Brox}

\maketitle

\begin{abstract}
Occlusions play an important role in disparity and optical flow estimation, since matching costs are not available in occluded areas and occlusions indicate depth or motion boundaries. Moreover, occlusions are relevant for motion segmentation and scene flow estimation. In this paper, we present an efficient learning-based approach to estimate occlusion areas jointly with disparities or optical flow. The estimated occlusions and motion boundaries clearly improve over the state-of-the-art. Moreover, we present networks with state-of-the-art performance on the popular KITTI benchmark and good generic performance. Making use of the estimated occlusions, we also show improved results on motion segmentation and scene flow estimation. 
\end{abstract}

\renewcommand{\thefootnote}{\fnsymbol{footnote}}
\footnotetext[1]{equal contribution}

\section{Introduction}


When applying dense correspondences to higher level tasks, there is often the desire for additional information apart from the raw correspondences. The areas in one image that are occluded in the other image are important to get an indication of potentially unreliable estimates due to missing measurements. A typical approach to estimate occluded areas is by computing correspondences in both directions and verifying their consistency post-hoc. However, since occlusions and correspondences are mutually dependent~\cite{mirrorflow,determiningOcclusions} and the presence of occlusions already negatively influences the correspondence estimation itself, post-processing is suboptimal and leads to unreliable occlusion estimates. 

Another valuable extra information in disparity maps and flow fields are explicit depth and motion boundaries, respectively. Referring to the classic work of Black\&Jepson~\cite{Black2000},  
"motion boundaries may be useful for navigation, structure from motion, video compression, perceptual organization and object recognition". 

In this paper, we integrate occlusion estimation as well as depth or motion boundary estimation elegantly with a deep network for disparity or optical flow estimation based on FlowNet 2.0~\cite{flownet2} and provide these quantities explicitly as output. In contrast to many prior works, this leads to much improved occlusion and boundary estimates and much faster overall runtimes. We quantify this improvement directly by measuring the accuracy of the occlusions and motion boundaries. We also quantify the effect of this improved accuracy on motion segmentation.

Furthermore we improved on some details in the implementation of the disparity and optical flow estimation networks from~\cite{dispnet,flownet,flownet2}, which gives us state-of-the-art results on the KITTI benchmarks. Moreover, the networks show good generic performance on various datasets if we do not fine-tune them to a particular scenario. While these are smaller technical contributions, they are very relevant for applications of optical flow and disparity. Finally, with state-of-the-art optical flow, disparity and occlusion estimates in place, we put everything together to achieve good scene flow performance at a high frame-rate, using only 2D motion information.
Using our predicted occlusions as input, we present a network 
that learns to interpolate the occluded areas to avoid the erroneous or missing information when computing the motion compensated difference between two disparity maps for scene flow. 

\section{Related Work} 

\textbf{Optical flow estimation with CNNs.} 
Optical flow estimation based on deep learning was pioneered by Dosovitsky~\etal~\cite{flownet}, who presented an end-to-end trainable encoder-decoder network. The work has been improved by Ilg et al.~\cite{flownet2}, who introduced a stack of refinement networks. Ranjan and Black~\cite{spynet} focused on efficiency and proposed a much smaller network based on the coarse-to-fine principle. 
Sun~\etal~\cite{pwcnet} extended this idea by introducing correlations at the different pyramid levels. Their network termed PWC-Net currently achieves state-of-the-art results. The coarse-to-fine approach, however, comes with the well-known limitation that the flow for small, fast-moving objects cannot be estimated. While this does not much affect the average errors of benchmarks, small objects can be very important for decisions in application scenarios. 

%
%
%
%
%
%
%

\textbf{Disparity estimation with CNNs.}
For disparity estimation, Zbontar et al.~\cite{zbontar} were the first to present a Siamese CNN for matching patches. Post-processing with the traditional SGM method~\cite{semiglobalmatching} yielded disparity maps. Other approaches to augment SGM with CNNs were presented by \cite{efficientstereo,sgmnets}. The first end-to-end learning framework was presented by Mayer et al.~\cite{dispnet}. The network named DispNetC was derived from the FlowNetC of Dosovitskiy et al.~\cite{flownet} restricted to rectified stereo images. It includes a correlation layer that yields a cost volume, which is further processed by the network.
Kendall~\etal~\cite{gcnet} presented GC-Net, which uses 3D convolutions to process the cost volume also along the disparity dimension and by using a differentiable softargmin operation. 
Pang~\etal~\cite{crl} extended DispNetC by stacking a refinement network on top, similar to FlowNet 2.0~\cite{flownet2}, with the difference that the second network is posed in a residual setting. 
In this work we also make use of network stacks with up to three networks and use their residual refinement. 


\textbf{Occlusion Estimation.}
%
%
%
Occlusion and optical flow estimation mutually depend on each other and are thus a typical 
chicken-and-egg problem~\cite{mirrorflow,determiningOcclusions}.
Humayun~\etal~\cite{learningOcclusionRegions} determine occlusions post-hoc by training a classifier on a broad spectrum of visual features and precomputed optical flow. P\'{e}rez-R\'{u}a~\etal~\cite{determiningOcclusions} do not require a dense optical flow field, but motion candidates, which are used to determine if a ``plausible reconstruction'' exists. 
Many other methods try to estimate optical flow and occlusions jointly. Leordeanu~\etal~\cite{affine_s2d} train a classifier based on various features, including the current motion estimate and use it repeatedly during energy minimization of the flow. Sun~\etal~\cite{localLayering} make use of superpixels and local layering for an energy formulation that is optimized jointly for layers, optical flow and occlusions. The most recent work from Hur~\etal~\cite{mirrorflow} uses consistency between forward and backward flows of two images, by integrating a corresponding constraint into an energy formulation. 
%
%
Since occlusions are directly related to changes in depth~\cite{binstereo}, it was quite common to consider them explicitly in disparity estimation methods \cite{binstereo,Ishikawa00globaloptimization,streo_correspondence,disp_occ_variational}. 

In this paper, we show that training a network for occlusion estimation is clearly beneficial, especially if the trained network is combined with a network formulation of disparity or optical flow estimation. 
We do not try to disentangle the chicken-and-egg problem, but instead solve this problem using the joint training procedure. 

\textbf{Depth and motion boundary estimation.} 
In many energy minimization approaches, depth or motion boundary estimation is implicitly included in the form of robustness to outliers in the smoothness constraint. Typically, these boundaries are not made explicit. An exception is Black\&Fleet~\cite{Black2000}, who estimate translational motion together with motion boundaries. Motion boundaries are also explicit in layered motion segmentation approaches. Most of these assume a precomputed optical flow, and only few estimate the segmentation and the flow jointly~\cite{local_layering,segflow}. 
Leordeanu~\etal~\cite{leordeanu} introduced a method for combined optimization of a boundary detector that also covers motion boundaries, while most other approaches make use of an external image boundary detector~\cite{gpb,fastedge}. Sundberg~\etal~\cite{sundberg} use gPb~\cite{gpb} and LDOF~\cite{ldof} to compute motion differences between regions adjacent to image boundaries. 
Weinzaepfel~\etal~\cite{weinzaepfel:hal-01142653} use a structured random forest trained on appearance and motion cues. Lei~\etal~\cite{boundaryFlow} present a fully convolutional Siamese network that is trained on annotated video segmentation. Using only the video segmentation ground-truth for training, they are able to infer the motion of boundary points during inference. 
For disparity and depth boundaries, the problem is very similar and most of the above mentioned methods could be applied to disparities, too. 
Jia~\etal~\cite{depth_boundaries} infer depth boundaries from color and depth images with a Conditional Random Field. 
In this paper, we obtain depth and motion boundaries also by a joint training procedure and by joint refinement together with occlusions and disparity or flow. 

\textbf{Scene Flow Estimation.}
Scene flow estimation was popularized for the first time by the work of Vedula et al.~\cite{vedula} and was later dominated by variational methods~\cite{huguet} \cite{quiroga} \cite{wedel}. Vogel et al.~\cite{vogel} combined the task of scene flow estimation with superpixel segmentation using a piecewise rigid model for regularization.
Schuster et al.~\cite{scene_ff} proposed a variational approach to interpolate sparse scene flow estimates from sparse matches. 
Behl et al.~\cite{isf} proposed a 3D scene flow method, which exploits instance recognition and 3D geometry information to obtain improved performance in texture-less, reflective and fast moving regions.

In this paper, we investigate scene flow estimation based on estimating correspondences only, without the use of 3D geometry information. The only learning based approach in a similar setting was proposed by Mayer et al.~\cite{dispnet}, but did not perform similarly well.


%
%

\section{Network Architectures\label{sec:network_archs}}

\begin{figure}[!ht]
    \begin{center}
    \subfigure[Extension of FlowNet2 with occlusions and residual connections.\label{fig:arch_fwd}]{
        \includegraphics[width=0.85\textwidth]{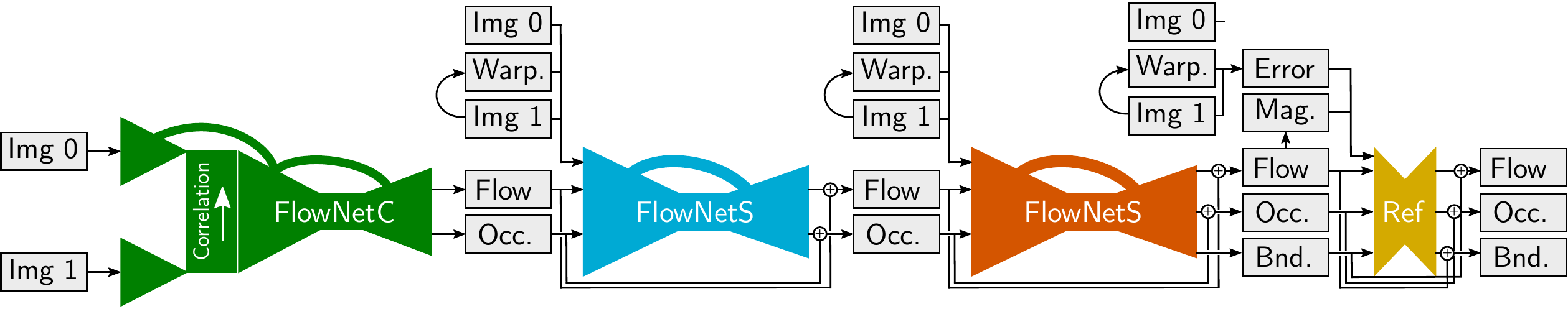}
    }
    
    \subfigure[Architecture for joint estimation of forward/backward flows and occlusions. See figure caption for symbol explanation.\label{fig:arch_joint}]{
        \includegraphics[width=0.7\textwidth]{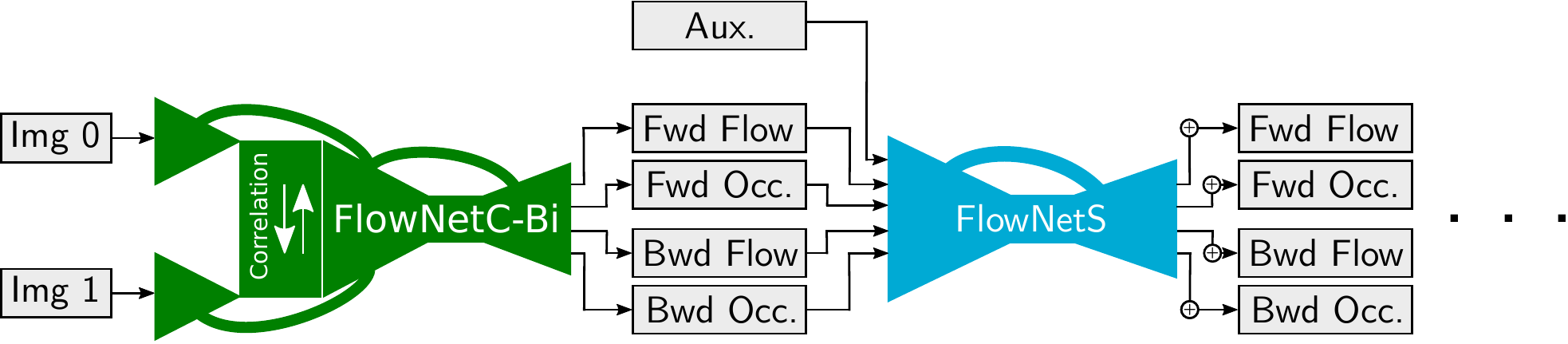}
    }
    
    \subfigure[Dual forward and backward estimation architecture with mutual warping. See figure caption for symbol explanation.\label{fig:arch_dual}]{
        \includegraphics[width=0.6\textwidth]{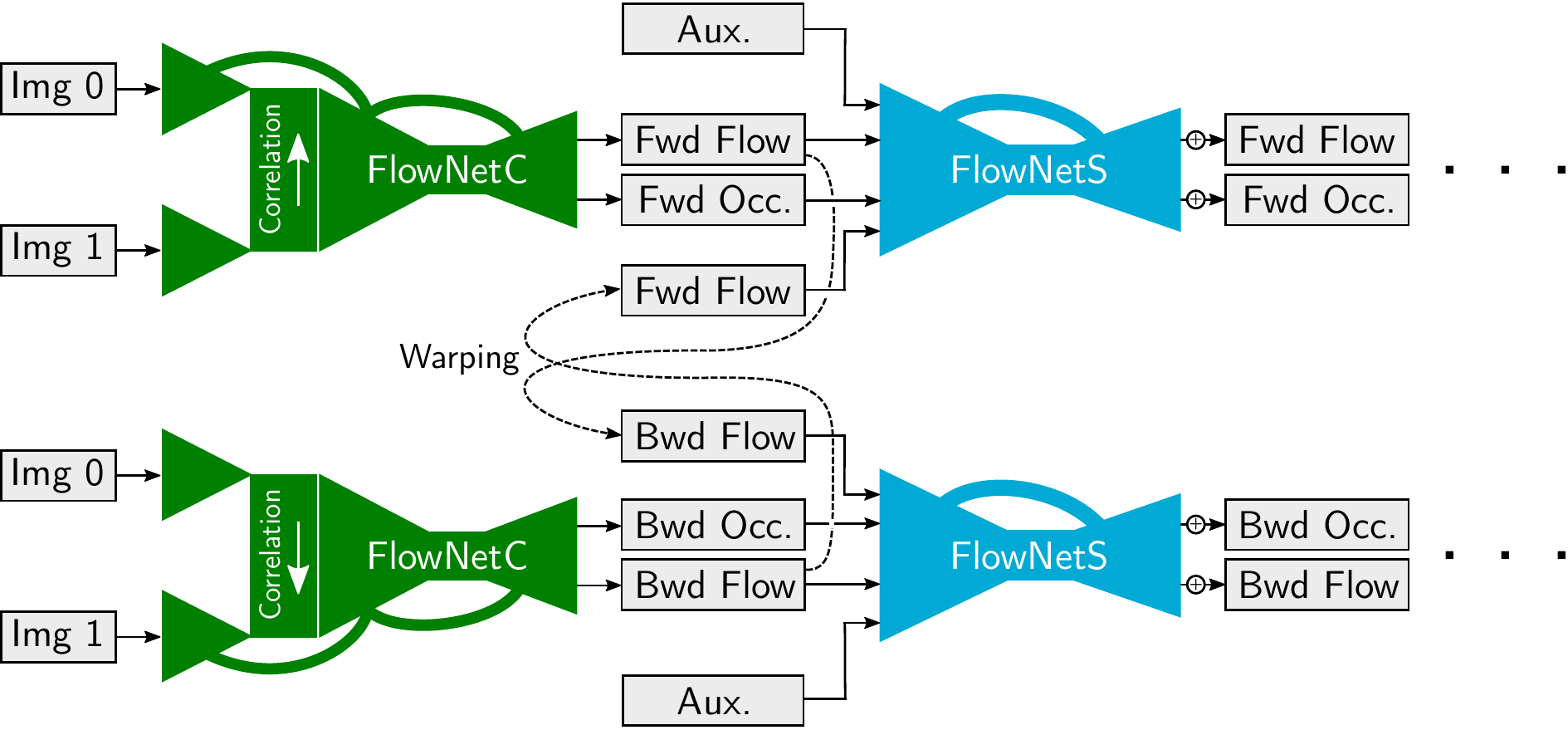}
    }

    \subfigure[Extending FlowNet-CSS and DispNet-CSS to a full scene flow network.\label{fig:arch_scene}]{
        \includegraphics[width=0.60\textwidth]{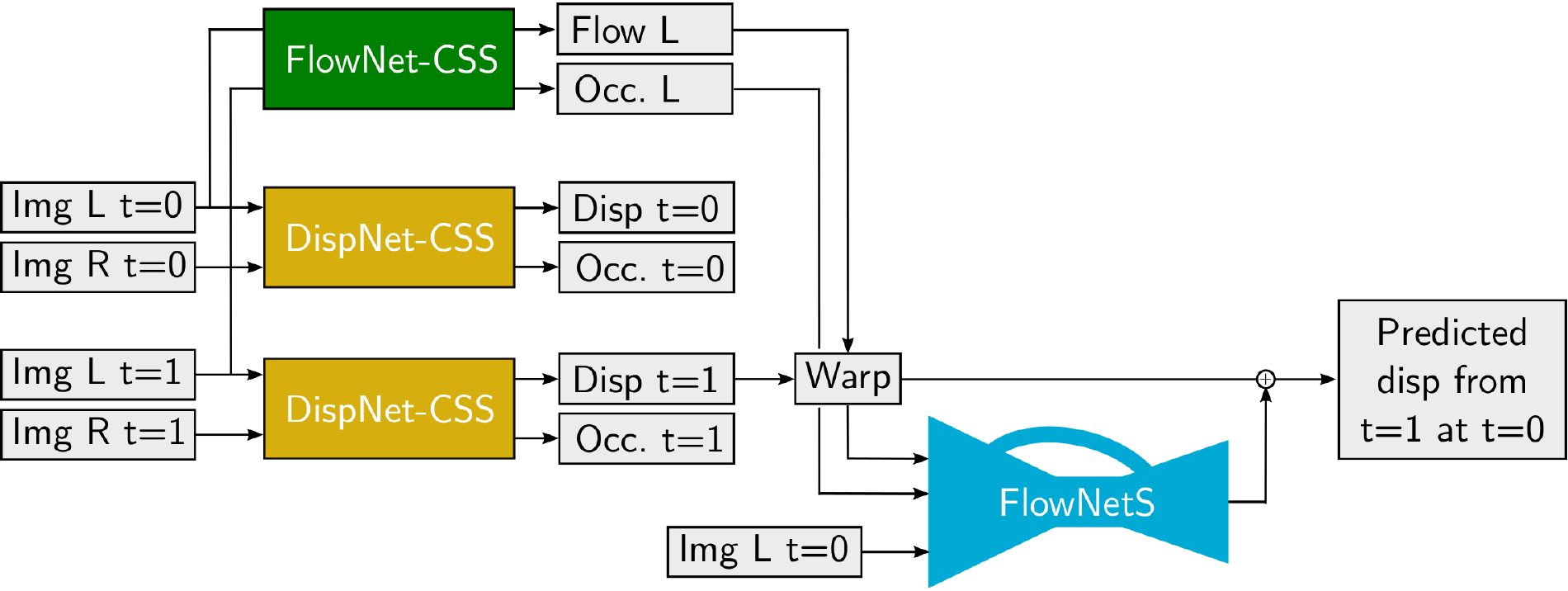}
    }
    \end{center}
    \caption{
    Overview of possible refinement stacks for flow, occlusions and motion boundaries. The residual connections are only shown in the first figure and indicated by + elsewhere. Aux. refers to the images plus a warped image for each input flow, respectively. 
    Architectures for the disparity case are analogous. 
    \label{fig:architectures}
    }
\end{figure}

We investigate estimating occlusions and depth or motion boundaries with CNNs together with disparity and optical flow. 
To this end, we build upon the convolutional encoder-decoder architectures from FlowNet~\cite{flownet} and the stacks from FlowNet 2.0~\cite{flownet2}. Our modifications are shown in Figure~\ref{fig:arch_fwd}. For simplicity, in the following we mention the flow case. The disparity case is analogous. 

In our version of~\cite{flownet2}, we leave away the small displacement network. In fact, the experiments from our re-implemented version show that the stack can perform well on small displacements without it. We still keep the former fusion network as it also performs smoothing and sharpening (see Figure~\ref{fig:arch_fwd}). We denote this network by the letter "R" in network names (e.g. FlowNet-CSSR). This network is only for refinement and does not see the second image. We further modify the stack by integrating the suggestion of Pang~\etal~\cite{crl} and add residual connections to the refinement networks. As in~\cite{flownet2}, we also input the warped images, but omit the brightness error inputs, as these can easily be computed by the network. 

Finally, we add the occlusions and depth or motion boundaries. 
While occlusions are important for refinement from the beginning, boundaries are only required in later refinement stages. Therefore we add the boundaries in the third network. Experimentally, we also found that when adding depth or motion boundary prediction in earlier networks, these networks predicted details better, but failed more rigorously in case of errors. Predicting exact boundaries early would be contrary to the concept of a refinement pipeline. 

Generally, in an occluded area, the forward flow from the first to the second image does not match the backward flow from the second to the first image. If the forward flow is correctly interpolated into the occluded regions, it resembles the flow of the background object. Since this object is not visible in the second image, the backward flow of the target location is from another object and forward and backward flows are inconsistent. Many classical methods use this fact to determine occlusions.

We bring this into the network architecture from Figure~\ref{fig:arch_joint}. In this version, we let the network estimate forward and backward flows and occlusions jointly. Therefore we modify FlowNetC to include a second correlation that takes a feature vector from the second image and computes the correlations to a neighborhood in the first image. We concatenate the outputs and also add a second skip connection for the second image. This setup is shown as FlowNetC-Bi in Figure ~\ref{fig:arch_joint}. Throughout the stack, we then estimate flow and occlusions in forward and backward directions. 

In the third variant from Figure~\ref{fig:arch_dual}, we model forward and backward flow estimation as separate streams and perform mutual warping to the other direction after each network. E.g., we warp the estimated backward flow after the first network to the coordinates of the first image using the forward flow. Subsequently we flip the sign of the warped flow, effectively turning it into a forward flow. The network then has the forward and the corresponding backward flow at the same pixel position as input. 

Finally, we use our networks for flow and disparity to build a scene flow extension.
For the scene flow task, the disparity at $t=0$ is required and the flow is extended by disparity change~\cite{dispnet} (resembling the change in the third coordinate). To compute this disparity change, one can estimate disparities at $t=1$, warp them to $t=0$ and compute the difference. However, the warping will be incorrect or undefined everywhere, where an occlusion is present. 
We therefore add the network shown in Figure~\ref{fig:arch_scene} to learn a meaningful interpolation for these areas given the warped disparity, the occlusions and the image.

\section{Experiments} 

\subsection{Training Data} 

For training our flow networks, we use the FlyingChairs~\cite{flownet}, FlyingThings3D~\cite{dispnet} and ChairsSDHom~\cite{flownet2} datasets. 
For training the disparity networks, we only use the FlyingThings3D~\cite{dispnet} dataset. 
These datasets do not provide the ground-truth required for our setting per-se. 
For FlyingChairs, using the code provided by the authors of~\cite{flownet}, we recreate the whole dataset including also backward flows, motion boundaries and occlusions. 
For FlyingThings3D, depth and motion boundaries are directly provided. We use flow and object IDs to determine occlusions. For ChairsSDHom, we compute motion boundaries by finding discontinuities among object IDs and in the flow, by using a flow magnitude difference threshold of $0.75$. To determine the ground-truth occlusions, we also use the flow and the object IDs. 

\subsection{Training Schedules and Settings}

For training our networks, we also follow the data and learning rate schedules of FlowNet 2.0~\cite{flownet2}. We train the stack network by network, always fixing the already trained networks.
Contrary to~\cite{flownet2}, for each step we only use half the number of iterations. The initial network is then a bit worse, but it turns out that the refinement can compensate for it well. We also find that the residual networks converge much faster. Therefore, we train each new network on the stack for $600k$ iterations on FlyingChairs and for $250k$ iterations on FlyingThings3D. Optionally, we follow the same fine-tuning procedure for small-displacements on ChairsSDHom as in~\cite{flownet2} (we add "-ft-sd" to the network names in this case). We use the caffe framework and the same settings as in~\cite{flownet2}, with one minor modification: we found that numerically scaling the ground-truth flow vectors (by a factor of $\frac{1}{20}$) yields noise for small displacements during optimization. We propose to change this factor to $1$. Since these are all minor modifications, we present details in the supplemental material.

To train for flow and disparity, we use the normal EPE loss. For small displacement training we also apply the suggested non-linearity of~\cite{flownet2}. To train for occlusions and depth or motion boundaries, we use a normal cross entropy loss with classes $0$ and $1$ applied to each pixel. To combine multiple losses of different kind, we balance their coefficients during the beginning of the training, such that their magnitudes are approximately equal. 

\subsection{Estimating Occlusions with CNNs} 

\begin{table}[t]
\centering
\resizebox{\textwidth}{!}{ 
\begin{tabular}{ |l|c|c|}
\hline
\multirow{2}{*}{Input}                                           & \multicolumn{2}{c|}{F-measure}  \\
\cline{2-3} 
                                                                 & FlyingThings3D~\cite{dispnet}  & Sintel clean~\cite{sintel}\\
\hline 
\hline 
Images 0+1                                                       &    $0.790$            & $0.545$ \\
\hline 
Images 0+1, GT fwd Flow                                          &    $0.932$            & $\textbf{0.653}$ \\
Images 0+1, GT fwd Flow, GT bwd flow                             &    $0.930$            & - \\
Images 0+1, GT fwd Flow, GT bwd flow warped+flipped              &    $\textbf{0.943}$   & - \\
Images 0+1, GT fwd Flow, GT fwd/bwd consistency                  &    $\textbf{0.943}$   & - \\
\hline
\end{tabular}
}
\vspace*{3mm}
\caption{Training a FlowNetS to estimate occluded regions from different inputs. Since Sintel~\cite{sintel} does not provide the ground-truth backward flow, we additionally report numbers on FlyingThings3D~\cite{dispnet}. The results show that contrary to literature~\cite{determiningOcclusions,affine_s2d,learningOcclusionRegions}, occlusion estimation is even possible from just the two images. Providing the optical flow, too, clearly improves the results
\label{tab:occ_gt_exp}
}
\end{table}

We first ran some basic experiments on estimating occlusions with a FlowNetS architecture and the described ground-truth data. In the past, occlusion estimation was closely coupled with optical flow estimation and in the literature is stated as "notoriously difficult"~\cite{affine_s2d} and a chicken-and-egg problem~\cite{mirrorflow,determiningOcclusions}. 
However, before we come to joint estimation of occlusions and disparity or optical flow, we start with a network that estimates occlusions independently of the optical flow or with optical flow being provided as input.

In the most basic case, we only provide the two input images to the network and no optical flow, i.e., the network must figure out by itself on how to use the relationship between the two images to detect occluded areas. As a next step, we additionally provide the ground-truth forward optical flow to the network, to see if a network is able to use flow information to find occluded areas. Since a classical way to detect occlusions is by checking the consistency between the forward and the backward flow as mentioned in Section~\ref{sec:network_archs}, we provide different versions of the backward flow: 1.) backward flow directly; 2.) using the forward flow to warp the backward flow to the first image and flipping its sign (effectively turning the backward flow into a forward flow up to the occluded areas); 3.) providing the magnitude of the sum of forward and backward flows, i.e., the classical approach to detect occlusions. 
From the results of these experiments in Table~\ref{tab:occ_gt_exp}, we conclude:

\textbf{Occlusion estimation without optical flow is possible.} 
In contrast to existing literature, where classifiers are always trained with flow input~\cite{determiningOcclusions,affine_s2d,learningOcclusionRegions,leordeanu} or occlusions are estimated jointly with optical flow~\cite{localLayering,mirrorflow}, we show that a deep network can learn to estimate the occlusions directly from two images. 

\textbf{Using the flow as input helps.} 
The flow provides the solution for correspondences and the network uses these correspondences. Clearly, this helps, particularly since we provided the correct optical flow. 

\textbf{Adding the backward flow marginally improves results.}
Providing the backward flow directly does not help. This can be expected, because the information for a pixel of the backward flow is stored at the target location of the forward flow and a look-up is difficult for a network to perform. 
Warping the backward flow or providing the forward/backward consistency helps a little.



\subsection{Joint Estimation of Occlusions and Optical Flow} 

\subsubsection{Within a single network.} 

In this section we investigate estimating occlusions jointly with optical flow, as many classical methods try to do~\cite{localLayering,mirrorflow}. Here, we provide only the image pair and therefore can use a FlowNetC instead of 
a FlowNetS. The first row of Table~\ref{tab:joint_est_flow_occ} shows that just occlusion estimation with a FlowNetC performs 
similar to the FlowNetS of the last section. Surprisingly, from rows one to three of Table~\ref{tab:joint_est_flow_occ} we find that joint flow estimation neither improves nor deproves the flow or the occlusion quality. In row four of the table we additionally   estimate the backward flow to enable the network to reason about forward/backward consistency. However, we find that this also does not affect performance much. 

When finding correspondences, occlusions need to be regarded by deciding that no correspondence exists for an occluded pixel and by  filling the occlusion area with some value inferred from the surroundings. Therefore, knowledge about occlusions is mandatory for correspondence and correct flow estimation. 
Since making the occlusion estimation in our network explicit does not change the result, we conclude that 
an end-to-end trained network for only flow 
already implicitly performs all necessary occlusion reasoning. By making it explicit, we obtain the occlusions as an additional output at no cost, but the flow itself remains unaffected. 

\begin{table}[t]
\centering
\begin{tabular}{|l|c|c|}
\hline
Configuration                                          & EPE                 & F-measure                      \\
\hline 
\hline 
FlowNetC estimating flow                               & $3.21$             & -                              \\
FlowNetC estimating occlusions                         & -                  & $\mathbf{0.546}$               \\
FlowNetC estimating flow + occlusions                  & $\mathbf{3.20}$    & $0.539$                        \\
FlowNetC-Bi estimating fwd/bwd flow and fwd occlusions & $3.26$             & $0.542$                        \\
\hline
\end{tabular}
\vspace*{4mm}
\caption{Joint estimation of flow and occlusions with a FlowNetC from Sintel train clean. 
Estimating occlusions neither improves nor degrades flow performance
\label{tab:joint_est_flow_occ}
}
\end{table}

\begin{table}[t]
\centering
\begin{tabular}{|l|c|c| }
\hline
Configuration                                        &  EPE                  & F-measure             \\
\hline 
\hline
Only flow as in FlowNet2-CS~\cite{flownet2}          & $2.28$                  &  -                  \\
+ occlusions (Figure~\ref{fig:arch_fwd})             & $\mathbf{2.25}$         &  $\mathbf{0.590}$   \\
+ bwd direction (Figure~\ref{fig:arch_joint})        & $2.77$                  &  $0.572$            \\
+ mutual warping (Figure~\ref{fig:arch_dual})        & $2.25$                  &  $0.589$            \\
\hline 
\end{tabular}
\vspace*{4mm}
\caption{Results of refinement stacks on Sintel train clean. Simply adding occlusions in a straightforward manner performs better or similar to more complicated approaches. In general, adding occlusions does not perform better than estimating only flow
\label{tab:occ_flow_refinement}
}
\end{table}

\subsection{With a refinement network}

In the last section we investigated the joint estimation of flow and occlusions, 
which in the literature is referred to as a "chicken-and-egg" problem. 
With our first network already estimating flow and occlusions, 
we investigate if the estimated occlusions can help refine the flow ("if a chicken can come from an egg").

To this end, we investigate the three proposed architectures from Section~\ref{tab:joint_est_flow_occ}. 
We show the results of the three variants in Table~\ref{tab:occ_flow_refinement}. While the architectures from Figures~\ref{fig:arch_fwd} and~\ref{fig:arch_dual} are indifferent about the additional occlusion input, the architecture with joint forward and backward estimation performs worse. 

Overall, we find that providing explicit occlusion estimates to the refinement does not help compared to estimating just the optical flow.
This means, either the occluded areas are already filled correctly by the base network, or in a stack without explicit occlusion estimation, the second network can easily recover occlusions from the flow and does not require the explicit input. 

We finally conclude that occlusions can be obtained at no extra cost, but do not actually influence the flow estimation, and that it is best to leave the inner workings to the optimization by using only the baseline variant (Figure~\ref{fig:arch_fwd}). This is contrary to the findings from classical methods.


\subsection{Comparing occlusion estimation to other methods \label{sec:occ_eval}} 

\begin{figure}[t]
\centering
    \begin{tabular}{cccc}
      \labelimage{0.23}{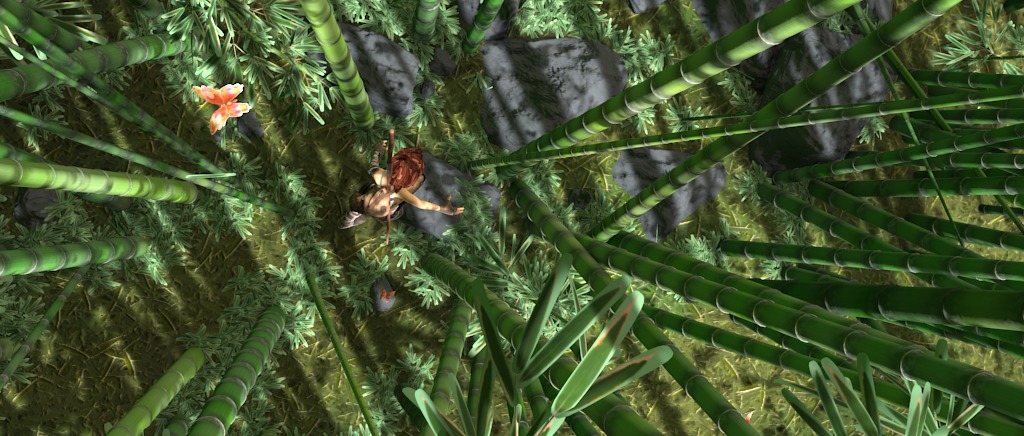}{Image 0}  
    & \labelimage{0.23}{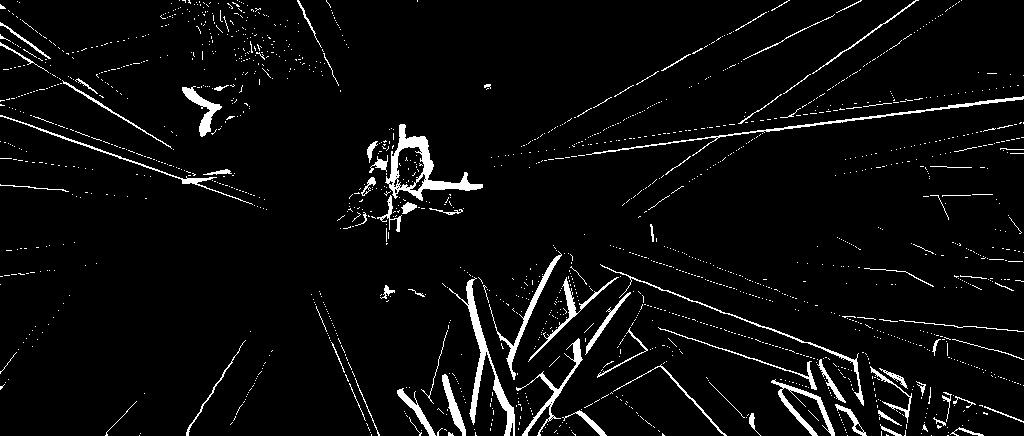}{Occ GT} 
    & \labelimage{0.23}{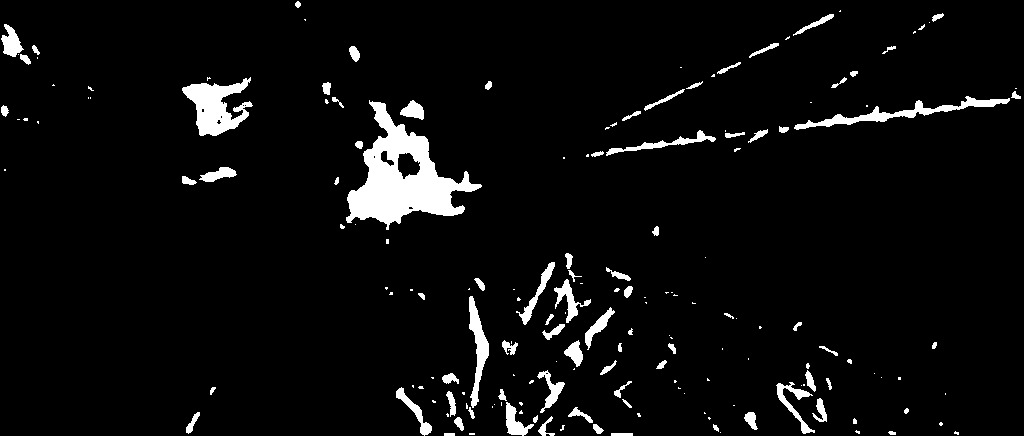}{S2DFlow ~\cite{affine_s2d}}
    & \labelimage{0.23}{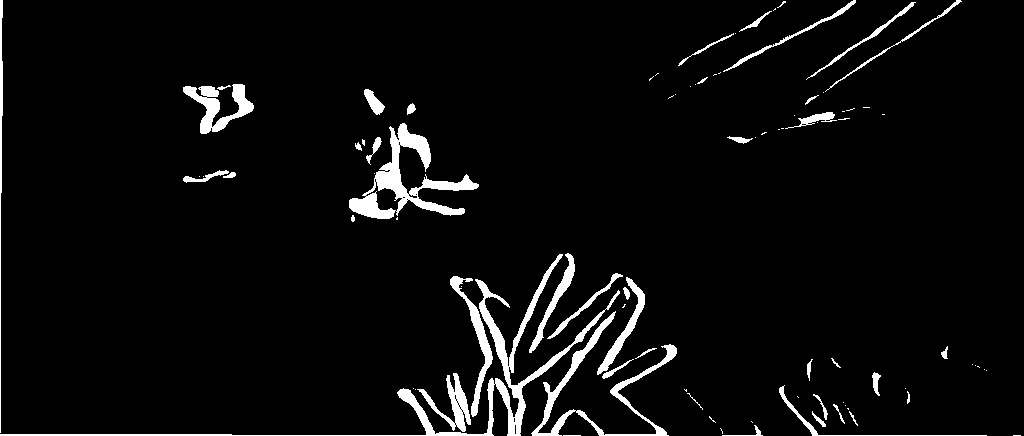}{Consistency}
     \\
      \labelimage{0.23}{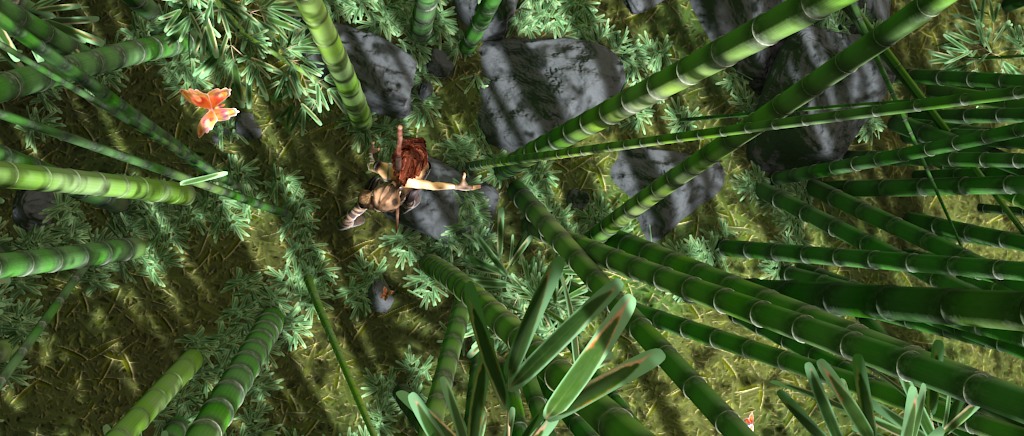}{Image 1} 
    & \labelimage{0.23}{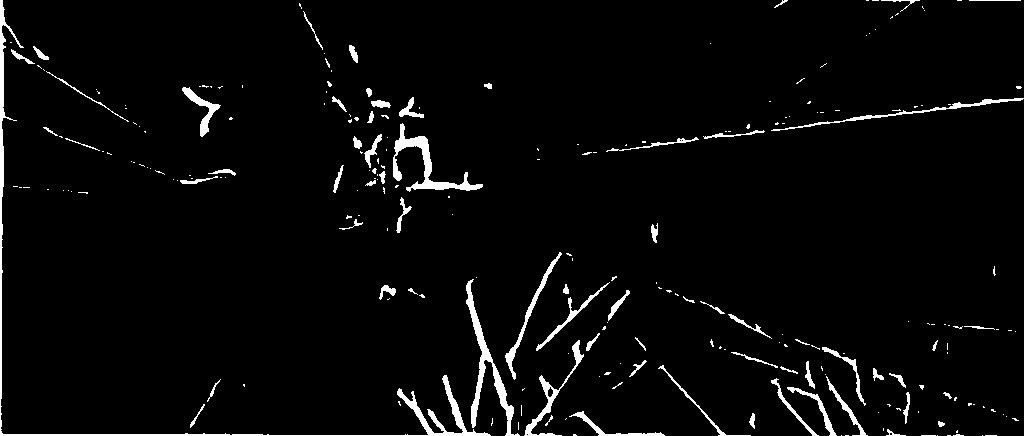}{Ours } 
    & \labelimage{0.23}{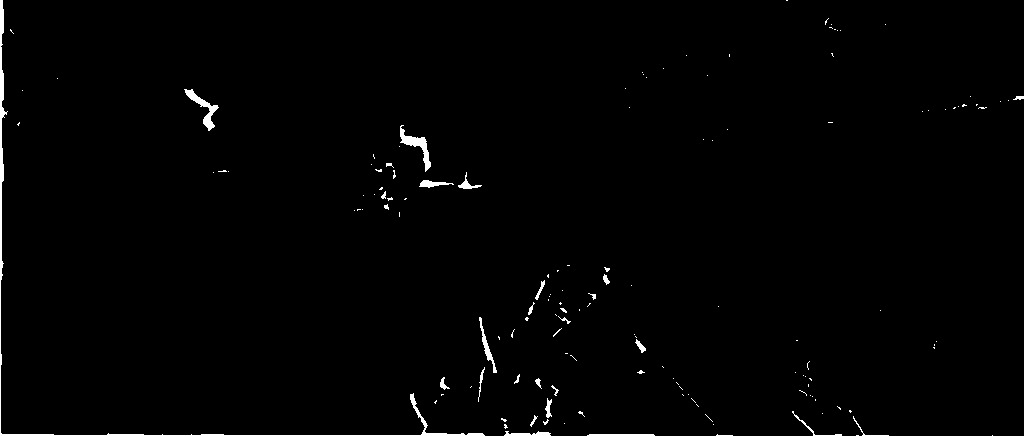}{MirrorFlow~\cite{mirrorflow}}  
    &
    \\
      \labelimage{0.23}{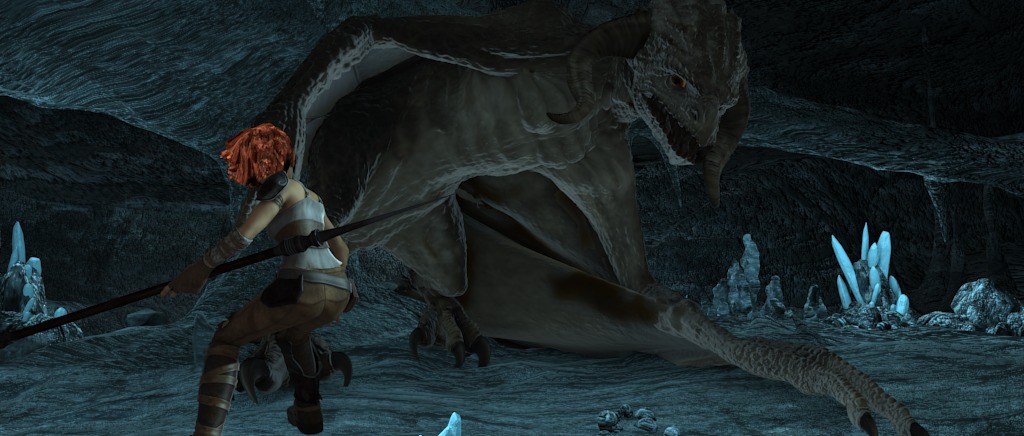}{Image 0}  
    & \labelimage{0.23}{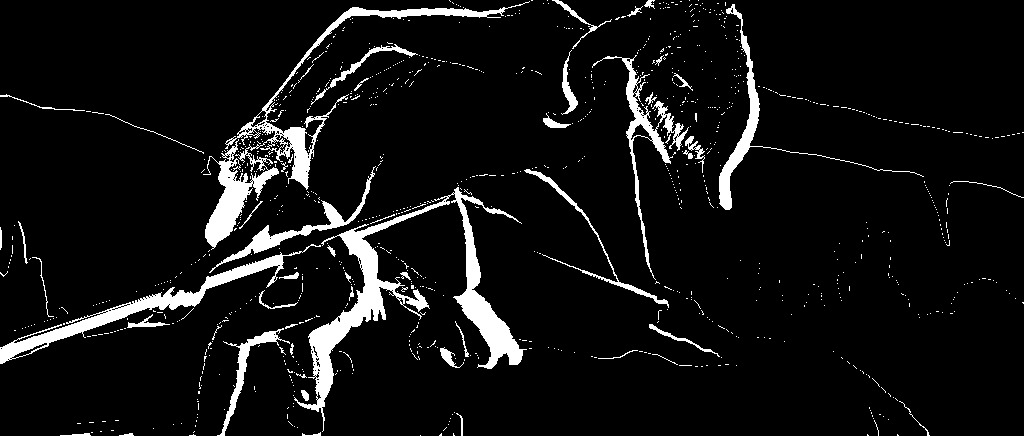}{Occ GT} 
    & \labelimage{0.23}{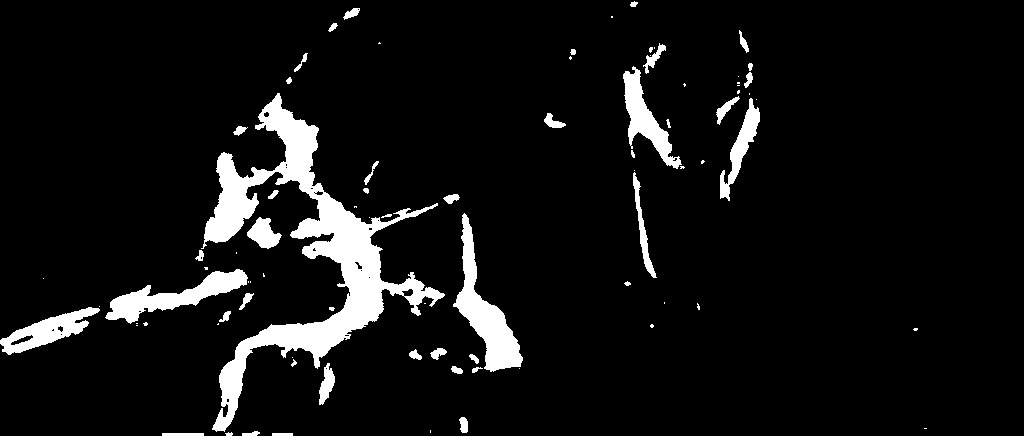}{S2DFlow ~\cite{affine_s2d}}
    & \labelimage{0.23}{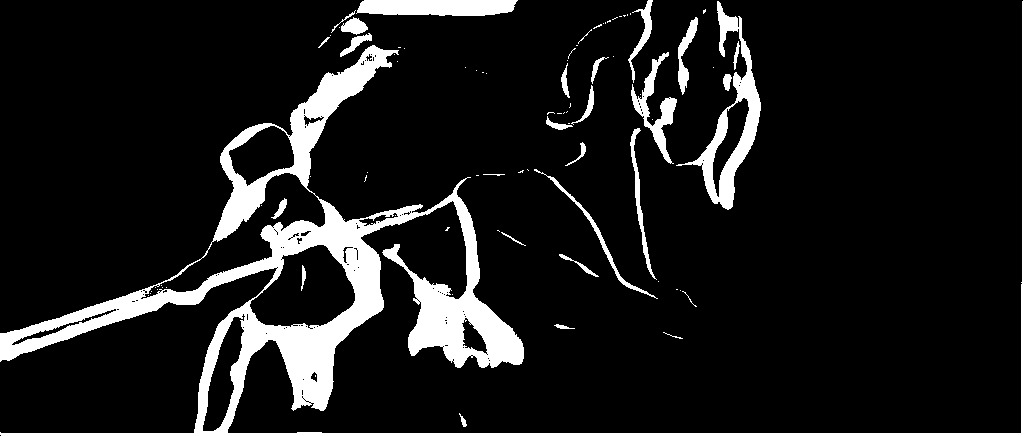}{Consistency}
     \\
      \labelimage{0.23}{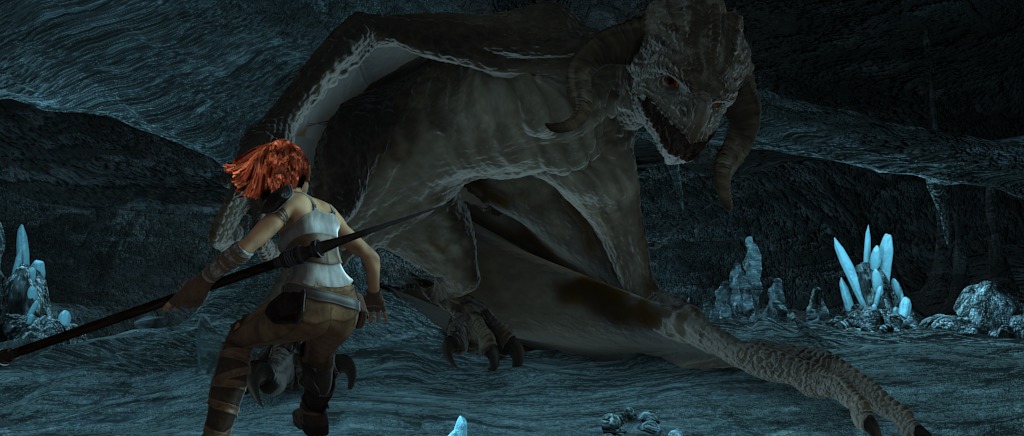}{Image 1} 
    & \labelimage{0.23}{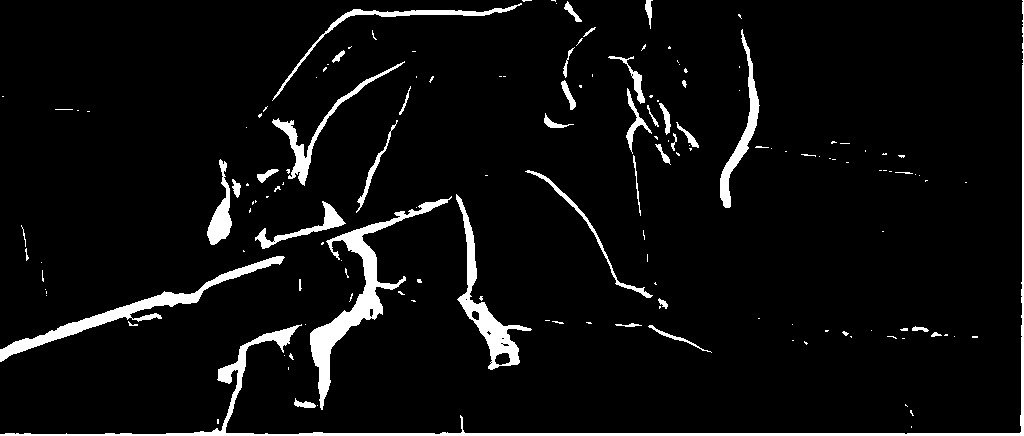}{Ours } 
    & \labelimage{0.23}{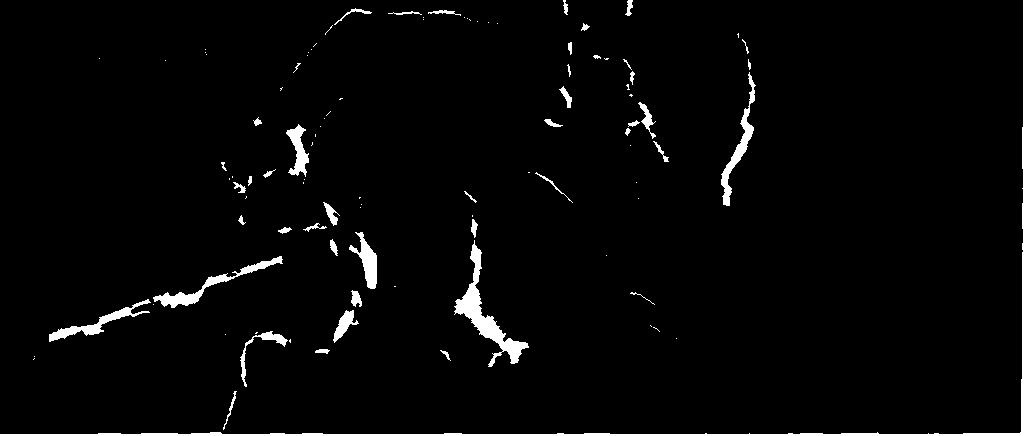}{MirrorFlow~\cite{mirrorflow}} 
    &
    \\
      \labelimage{0.23}{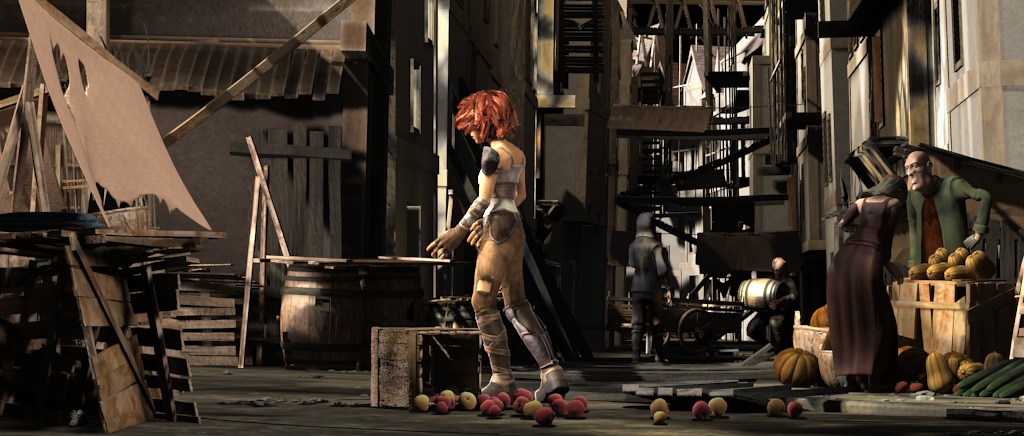}{Image 0}  
    & \labelimage{0.23}{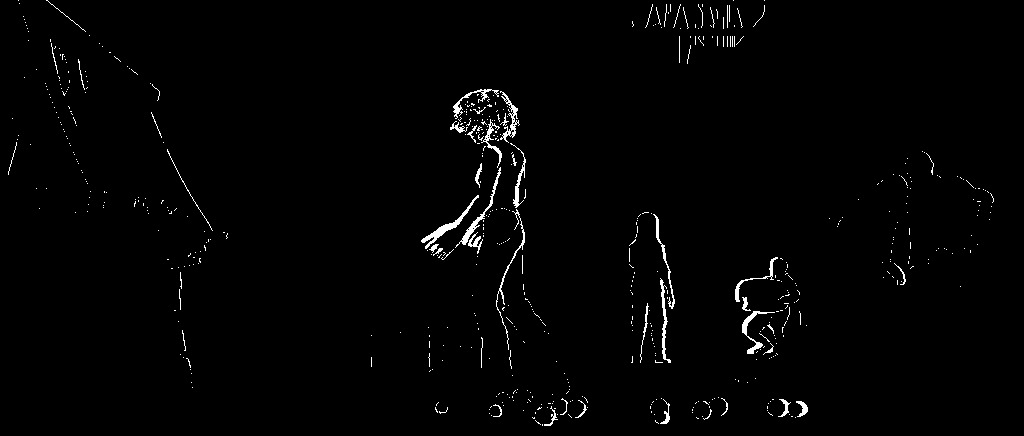}{Occ GT} 
    & \labelimage{0.23}{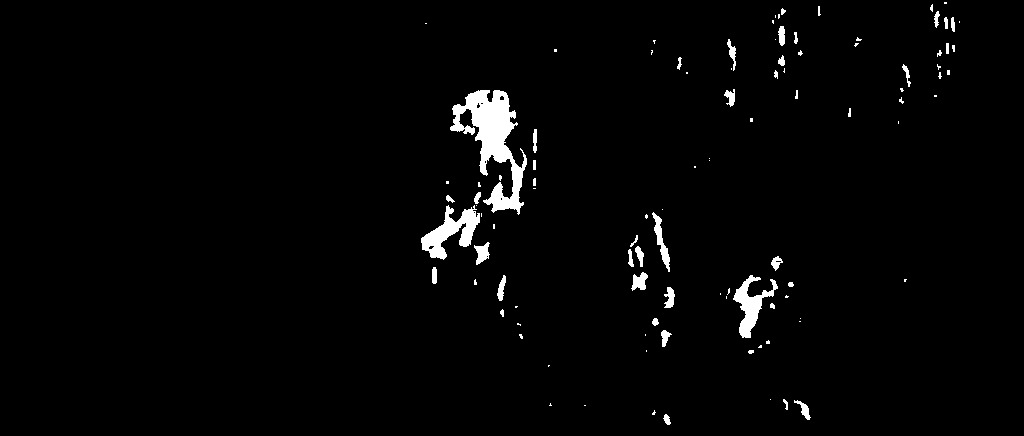}{S2DFlow ~\cite{affine_s2d}}
    & \labelimage{0.23}{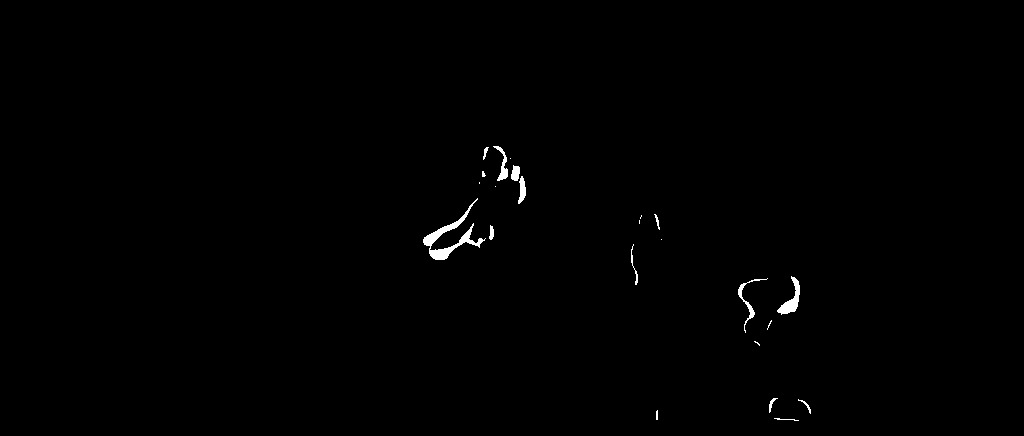}{Consistency}
     \\
      \labelimage{0.23}{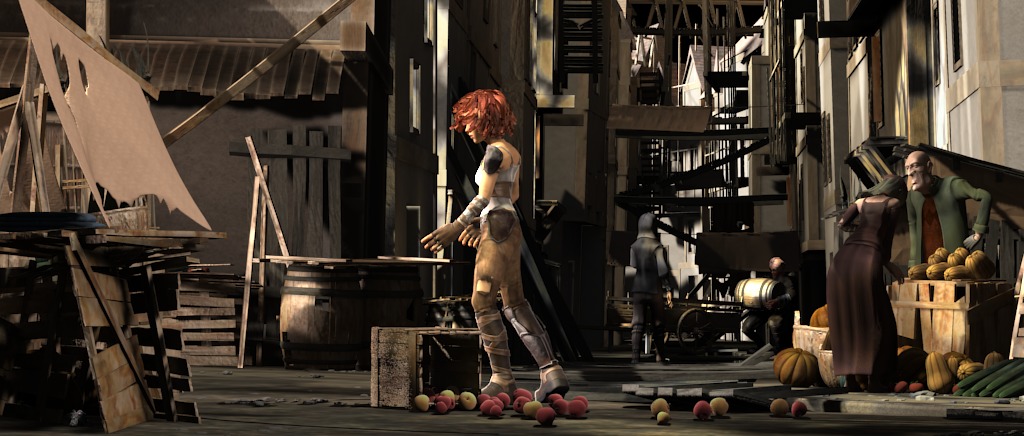}{Image 1} 
    & \labelimage{0.23}{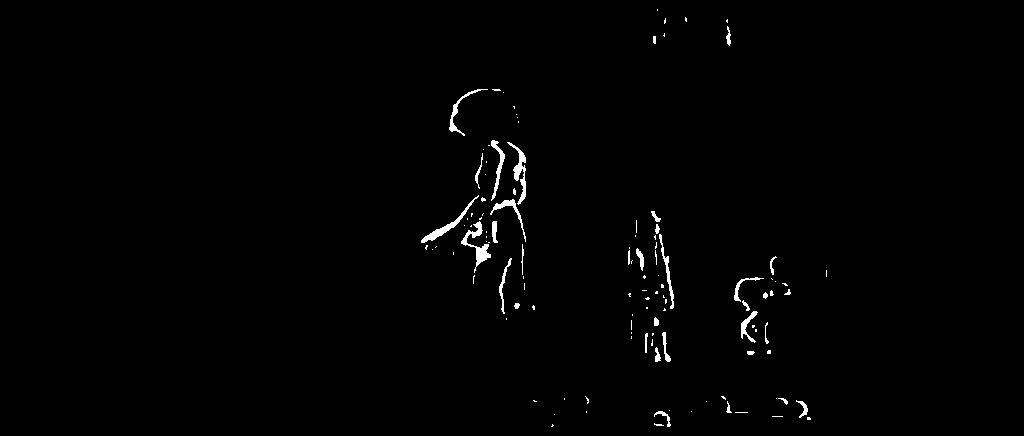}{Ours } 
    & \labelimage{0.23}{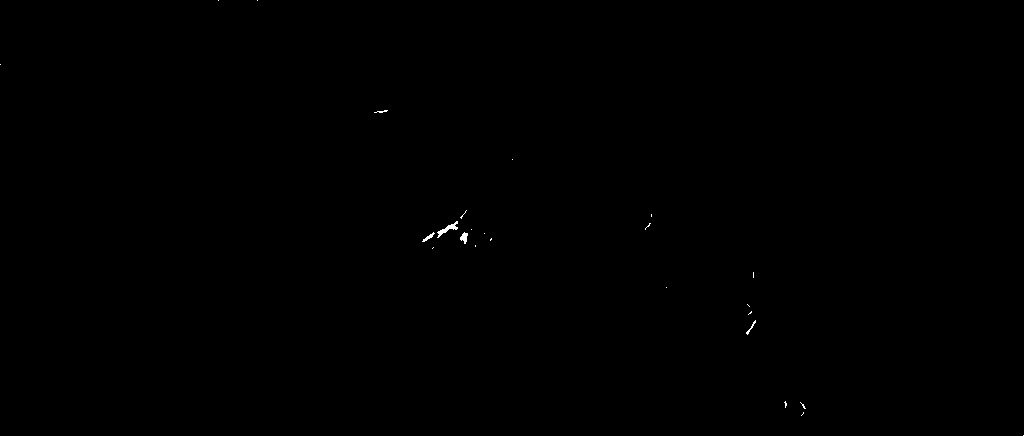}{MirrorFlow~\cite{mirrorflow}}
    &
    \\
    \end{tabular}  
\caption{
         Qualitative results for occlusions. In comparison to other methods and the forward-backward consistency check, our method is able to capture very fine details.         
     \label{fig:qual_occ}
        } 
\end{figure}

In Tables~\ref{tab:occ_tab_disp} and~\ref{tab:occ_tab_flow} we compare our occlusion estimations to other methods. 
For disparity our method outperforms Kolmogorov et al.~\cite{kz2} for all except one scene. 
For the more difficult case of optical flow, we outperform all existing methods by far. 
This shows that the chicken-and-egg problem of occlusion estimation is much easier to handle 
with a CNN than with classical approaches~\cite{disp_occ_variational,kz2,mirrorflow,affine_s2d} and that CNNs can perform very well at occlusion reasoning. This is confirmed by the qualitative results of Figure~\ref{fig:qual_occ}. While consistency checking is able to capture mainly large occlusion areas, S2DFlow~\cite{affine_s2d} also manages to find some details. MirrorFlow~\cite{mirrorflow} in many cases misses details. Our CNN on the other hand is able to estimate most of the fine details. 

\begin{table}[t]
\centering

\begin{tabular}{|l|c|c|c|c||c|c|}
\hline
\multirow{2}{*}{Method}                      & \multicolumn{6}{c|}{F-Measure} \\
\cline{2-7}
                                          & Cones     & Teddy         & Tsukuba & Venus & Sintel clean & Sintel final\\
\hline
\hline
Kolmogorov ~\etal \cite{kz2}         
                                 & ${0.45}$  & $\mathbf{0.63}$   & ${0.60}$     & ${0.41}$  &    - & -  \\
Tan  ~\etal \cite{disp_occ_variational}  
                                  & $0.44$  & $0.40$   & $0.50$     & $0.33$  &    -  & -   \\
\hline
Ours                               & $\mathbf{0.91}$  & ${0.57}$   & $\mathbf{0.68}$    & $\mathbf{0.44}$   &    $\mathbf{0.76}$ & $\mathbf{0.72}$ \\  
\hline 
\end{tabular}
\vspace*{5mm}
\caption{Comparison of estimated disparity occlusions from our DispNet-CSS to other methods on examples from the Middlebury 2001 and 2003 datasets (results of Kolmogorov et al.~\cite{kz2} and Tan et al.~\cite{disp_occ_variational} taken from~\cite{disp_occ_variational}) and the Sintel train dataset.
Only in the scene Teddy of Middlebury our occlusions are outperformed by Kolmogorov~\etal~\cite{kz2}
\label{tab:occ_tab_disp}
}
\end{table}

\begin{table}[t]
\centering
\begin{tabular}{|l|c|c|c|}
\hline
\multirow{2}{*}{Method} & 
\multirow{2}{*}{Type} &
\multicolumn{2}{c|}{F-Measure} 
\\
\cline{3-4}
 & 
 &
clean &
final 
\\
\hline
\hline
FlowNet2~\cite{flownet2} & 
consistency &
$0.377$ & $0.348$
\\ 
MirrorFlow~\cite{mirrorflow} &
estimated &
$0.390$ &
$0.348$
\\
S2DFlow~\cite{affine_s2d} & 
estimated &
${0.470}$ & 
${0.403}$
\\
\hline 
Ours   & 
estimated &
$\mathbf{0.703}$ & 
$\mathbf{0.654}$
\\ 
\hline
\end{tabular}

\vspace*{5mm}
\caption{Comparison of the occlusions from FlowNet-CSSR-ft-sd to other occlusion estimation methods on the Sintel train dataset.
For the first entry, occlusions were computed using forward/backward consistency post-hoc. 
The proposed approach yields much better occlusions 
\label{tab:occ_tab_flow}
}
\end{table}

\subsection{Motion Boundary Estimation} 

\begin{figure}[ht]

\centering
    \subfigure[Image 0\label{fig:mb_img0}]{
        \includegraphics[width=0.31\textwidth]{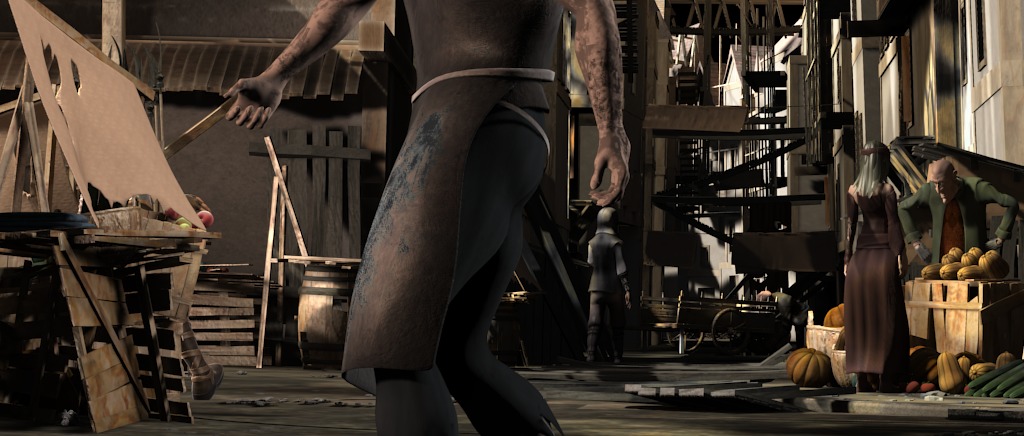}
    }
    \subfigure[Boundary ground-truth\label{fig:mb_gt}]{
        \includegraphics[width=0.31\textwidth]{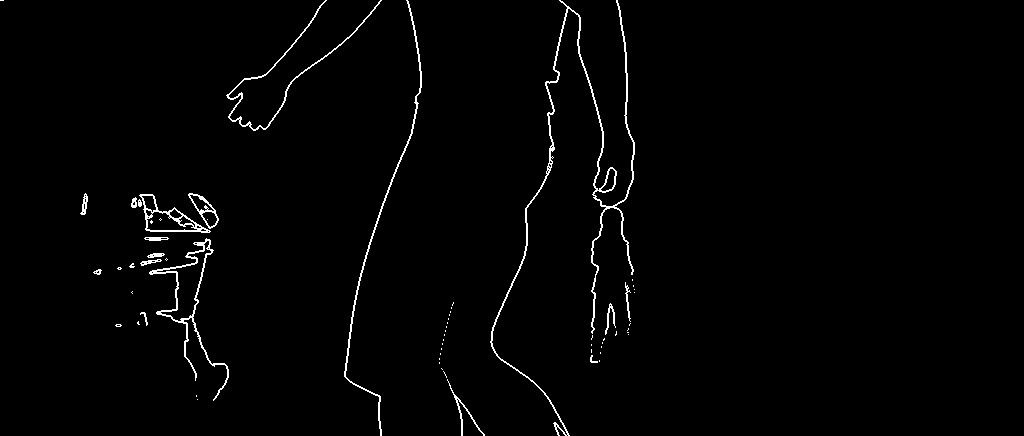}
    }
    \subfigure[Weinzaepfel~\etal~\cite{weinzaepfel:hal-01142653}]{
        \includegraphics[width=0.31\textwidth]{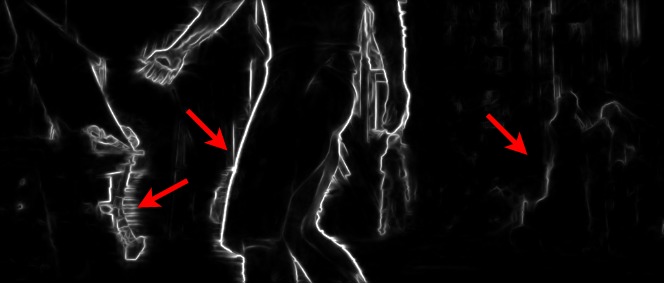}
    }

    \subfigure[Flow ground-truth \label{fig:flow_gt}]{
        \includegraphics[width=0.31\textwidth]{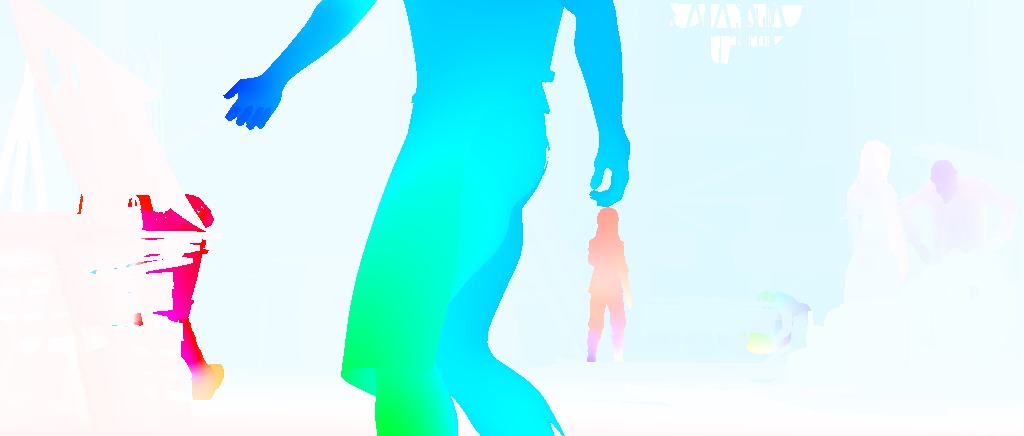}
    }
    \subfigure[Ours (hard)]{
        \includegraphics[width=0.31\textwidth]{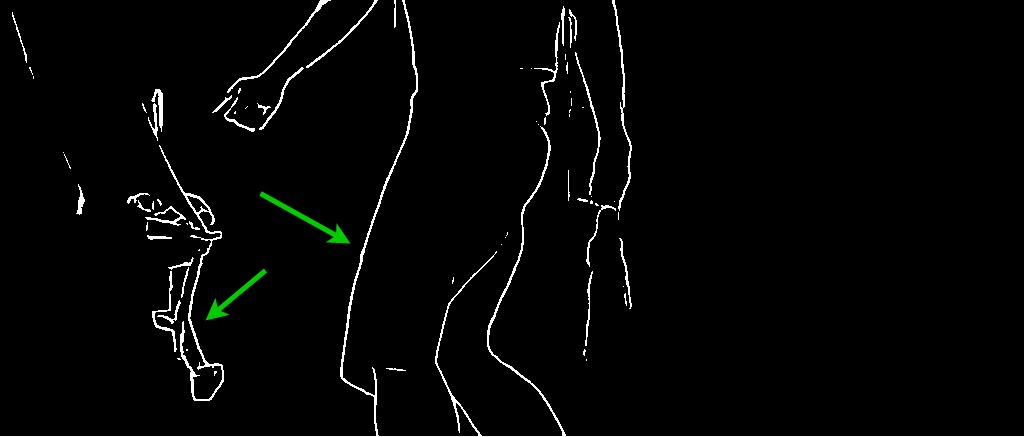}
    }
    \subfigure[Ours(soft)]{
        \includegraphics[width=0.31\textwidth]{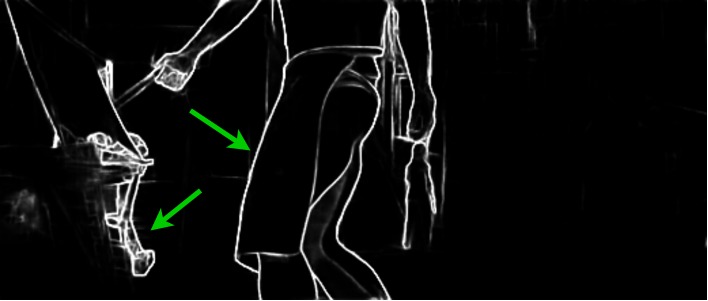}
    }
    \caption{
        Motion boundaries on Sintel train clean.  
        Our approach succeeds to detect the object in the background and has less noise 
        around motion edges than existing approaches (see green arrows). Weinzaepfel et al. detect some correct motion details in the background. However, these details are not captured in the ground-truth. 
    \label{fig:qual_bnd}
    }
\end{figure}

For motion boundary estimation we compare to Weinzaepfel~\etal~\cite{weinzaepfel:hal-01142653}, 
which is to the best of our knowledge the best available method. It uses a random forest classifier and
 is trained on the Sintel dataset. Although we do not train 
on Sintel, from the results of Table~\ref{tab:results_motion_boundaries}, our CNN outperforms 
their method by a large margin. The improvement in quality is also very well visible from 
the qualitative results from Figure~\ref{fig:qual_bnd}. 

\subsection{Application to Motion Segmentation} 

We apply the estimated occlusions to the motion segmentation framework from Keuper~\etal~\cite{keuper15a}. 
This approach, like \cite{Bro10c},
computes long-term point trajectories based on optical flow.
For deciding when a trajectory ends, the method depends on reliable occlusion estimates. 
These are commonly computed using the post-hoc consistency of forward and backward flow, which was shown to perform badly in Section~\ref{sec:occ_eval}. 
We replace the occlusion estimation with the occlusions from our FlowNet-CSS. 
Table~\ref{tab:results_motion_seg} shows the clear improvements obtained by the more reliable occlusion estimates on the common FBMS-59 motion segmentation benchmark. In row four, we show how adding our occlusions to flow estimations of FlowNet2 can 
improve results. This shows that by only adding occlusions, we recover 30 objects instead of 26. The last result from our flow and occlusions together further improve the results. Besides the direct quantitative and qualitative evaluation from the last sections, this shows the usefulness of our occlusion estimates in a relevant application. Our final results can produce results that are even better than the ones generated by the recently proposed third order motion segmentation with multicuts~\cite{Keuper_2017_ICCV}. 


\begin{table}[t]
\centering 
    \begin{tabular}{|l|c|c|}
    \hline
         Method & Sintel clean & Sintel final \\
    \hline
    \hline
        Weinzaepfel~\etal~\cite{weinzaepfel:hal-01142653}  & 76.3 & 68.5 \\
    \hline 
        Ours & $\mathbf{86.3}$ & $\mathbf{79.5}$ \\
    \hline
    \end{tabular}
\vspace*{4mm}
 \caption{
        Comparison of our motion boundary estimation to Weinzaepfel et al.~\cite{weinzaepfel:hal-01142653} on the Sintel train dataset. 
        The table shows the mean average precision computed with their evaluation code.
        Although Weinzaepfel et al.~\cite{weinzaepfel:hal-01142653} trained on Sintel train clean, our method outperforms theirs by a large margin
\label{tab:results_motion_boundaries}
    }
\end{table}

\begin{table}[t!]
\centering 
  \begin{tabular}{|l|cccc|}
    \hline 
    \multirow{2}{*}{Method} & \multicolumn{4}{c|}{FBMS test set (30 sequences)} \\[0.5mm]
           & {~Precision~} & {~Recall~} & {~F-Measure~} &  {~\#Objects~} \\
    \hline 
    \hline 
    {Third Order Multicut \cite{Keuper_2017_ICCV}} & 87.77\% & 71.96\% & 79.08\%  &  {29/69} \\
    \hline 
    {DeepFlow \cite{weinzaepfel:hal-00873592}} & 88.20\% & {69.39\%} & { 77.67\%}  &  {26/69} \\
    {FlowNet2} & {86.73\%} & {68.77\%} & {76.71\%}  & {26/69}\\
    {FlowNet2 + our occ} & {85.67\%} & {70.15\%} & {77.14\%}  & {30/69}\\   
    {Ours} & {\bf 88.71\%} & {\bf 73.60\%} & {\bf 80.45\%} & {\bf 31/69}\\
    \hline
  \end{tabular}%
\vspace*{5mm}
 \caption{\label{tab:results_motion_seg} 
  Results of motion segmentation from Keuper~\etal~\cite{keuper15a} on the FBMS-59 test set~\cite{Bro10c,Ochs14} (with 
  sampling density 8px). 
  The fourth row uses flows from FlowNet2~\cite{flownet2} combined with our occlusions. The improved results show that 
  occlusions help the motion segmentation in general. The last row shows the segmentation using our flow and occlusions, 
  which performs best and also improves over the recent state-of-the-art on sparse motion segmentation using higher order motion models~\cite{Keuper_2017_ICCV}
  }
\end{table}

\subsection{Benchmark results for disparity, optical flow, and scene flow}

Finally, we show that besides the estimated occlusions and depth and motion boundaries, our disparities and optical flow achieve state-of-the-art performance. 
In Table~\ref{tab:compare_disp} we show results for the common disparity benchmarks. 
We also present smaller versions of our networks by scaling the number of channels in each layer down to $37.5\%$ as suggested in~\cite{flownet2} (denoted by css). While this small version yields a good speed/accuracy trade-off, the larger networks rank second on KITTI 2015 and are the top ranked methods on KITTI 2012 and Sintel. 

In Table~\ref{tab:compare_flow} we show the benchmark results for optical flow. We perform on-par on Sintel, while we set the new state-of-the-art on both KITTI datasets. 

In Table~\ref{tab:compare_scene} we report numbers on the KITTI 2015 scene flow benchmark.  The basic scene flow approach warps the next frame disparity maps into the current frame (see~\cite{sgm_ff}) using forward flow. Out of frame occluded pixels cannot be estimated this way. To mitigate this problem we train a CNN to reason about disparities in occluded regions (see the architecture from Figure~\ref{fig:arch_scene}). This yields clearly improved results that get close to the state-of-the-art while the approach is orders of magnitude faster.

\begin{table}[ht]
\centering 
\newcommand{\rowline} {\hhline{|~||~|~~~~||~|}} 
\newcommand{\headerline} {\hhline{|=||=|====||=|}} 
\newcommand{\headertitle}[1]{\textbf{#1} &&&&&&\\}

\begin{tabular}{|l||c|cccc||c|}
\hhline{|-||-|----||-|}
\textbf{Method} 
&\textbf{Sintel}
&\multicolumn{2}{c}{\textbf{KITTI}}
&\multicolumn{2}{c||}{\textbf{KITTI}}
& Runtime\\

& (clean)
&\multicolumn{2}{c}{(2012)}
&\multicolumn{2}{c||}{(2015)}
& 
\\
& AEE
& AEE & Out-noc
& AEE & D1-all
& 
\\
& \textit{train}  
& \textit{train} & \textcolor{black}{\textit{test}} 
& \textit{train} & \textcolor{black}{\textit{test}}
& \textit{(s) }
\\
\headerline 

\headertitle{Standard}
\headerline 
SGM ~\cite{semiglobalmatching}
& $19.62$    
& $10.06$ & -  
& $7.21$  & $10.86\%$   
& $1.1$
\\


\headerline 
\headertitle{CNN based}
\headerline 
DispNetC ~\cite{dispnet} 
& $\pz5.66$  
& $1.75$  & -  
& $1.59$  & -  
& $0.06$
\\
DispNetC-ft \cite{dispnet}
& $21.88$ 
& $1.48$ & $4.11\%$  
& $(0.68)$ & $4.34\%$  
& $0.06$
\\
CRL ~\cite{crl}
&$16.13$ 
& $1.11$    & -  
& $(0.52)$  & $2.67\%$	  
& $0.47$
\\
GC-Net \cite{gcnet}
& -               
& -  & $\mathbf{1.77\%}$  
& -  & $2.87\%$  
& $0.90$ 
\\
MC-CNN-acrt ~\cite{zbontar}
& -  
& -   & $2.43\%$  
& -      & $3.89\%$  
& $67$
\\
DRR \cite{drr}
& -  
& -  & -  
& -  & $3.16\%$	  
& $0.4$
\\
L-ResMatch	~\cite{lresmatch}
& -  
& -  & $2.27 \%$  
& -  & $3.42 \%$  
& $42$
\\


\headerline 
\headertitle{With joint occ. est.}
\headerline 
SPS stereo \cite{sps_stereo}
& -  
& -  & $3.39\%$    
& -  & $5.31\%$  
& $2$
\\

Our DispNet-CSS
& $\mathbf{2.33}$  
& $1.40$ & -  
& $1.37$ & -  
& $0.07$
\\
Our DispNet-CSS-ft
& $5.53$   
& $(0.72)$  & $1.82\%$  
& $(0.71)$  & $\mathbf{2.19\%}$  
& $0.07$
\\
Our DispNet-css 
& $2.95$    
& $1.53$ & -  
& $1.49$& -  
& $0.03$
\\


\hhline{|-||-|----||-|}

\end{tabular}


\vspace*{5mm}
\caption{
\label{tab:compare_disp}
Benchmark results for \textbf{disparity estimation}. 
We report the average endpoint error (AAE) for Sintel. On KITTI, Out-noc and D1-all are used for the benchmark ranking on KITTI 2012 and 2015, respectively. Out-noc shows the percentage of outliers with errors more than $3$px in non-occluded regions, whereas D1-all shows the percentage in all regions. 
Entries in parentheses denote methods that were finetuned on the evaluated dataset.
Our network denoted with "-ft" is finetuned on the respective training datasets.
We obtain state-of-the-art results on the Sintel and KITTI 2015. Also, our networks generalize well across domains, as shown by the good numbers of the non-finetuned networks and the reduced drop in performance for a network finetuned on KITTI and tested on Sintel
} 
\end{table}

\begin{table}[h]
\centering 
\newcommand{\rowline} {\hhline{|~||~~~~|~~~~||~|}} 
\newcommand{\headerline} {\hhline{|=||====|====||=|}} 

\resizebox{\textwidth}{!}{%
\begin{tabular}{|l||cccc|cccc||c|}
\hhline{|-||----|----||-|}
\textbf{Method} 
&\multicolumn{2}{c}{\textbf{Sintel}} 
&\multicolumn{2}{c|}{\textbf{Sintel}}
&\multicolumn{2}{c}{\textbf{KITTI}}
&\multicolumn{2}{c||}{\textbf{KITTI}}
& \textbf{Runtime} \\

&\multicolumn{2}{c}{(clean)} 
&\multicolumn{2}{c|}{(final)}
&\multicolumn{2}{c}{(2012)}
&\multicolumn{2}{c||}{(2015)}
& (s)
\\
\hhline{|~||~~~~|~~~~||~|}
&\multicolumn{2}{c}{AEE} 
&\multicolumn{2}{c|}{AEE} 
& \textit{AEE} & \textit{OUT-noc}  
& \textit{AEE} & \textit{F1-all} 
& 
\\
\hhline{|~||~~~~|~~~~||~|}
& \textit{train} & \textcolor{black}{\textit{test} }
& \textit{train} & \textcolor{black}{\textit{test} }
& \textit{train} & \textcolor{black}{\textit{test} }
& \textit{train} & \textcolor{black}{\textit{test} }
& 
\\
\headerline 

\textbf{Standard}
&&&&&&&&&\\
\headerline 

EpicFlow~\cite{epicflow}
& ${2.27}$  & $4.12$  
& ${3.56}$  & $6.29$  
& $\mathbf{3.09}$  & $7.88\%$ 
& ${9.27}$  & $26.29\%$	  
& $42$   
\\
\rowline 

FlowfieldsCNN~\cite{flowfieldscnn}
& -  & $3.78$ 
& -  & $5.36$  
& -  & ${4.89\%}$ 
& -  & $18.68\%$	  
& $23$  
\\
\rowline 

DCFlow~\cite{dcflow}
& -  & $\mathbf{3.54}$  
& -  & ${5.12}$ 
& -  & -  
& -  & ${14.86\%}$	  
& $\pz{9}$
\\

\rowline 

\headerline 
\textbf{CNN based}
&&&&&&&&& \\
\headerline 
FlowNet2~\cite{flownet2}
& $\mathbf{2.02}$  & ${3.96}$  
& $\mathbf{3.14}$  & $6.02$  
& ${4.09}$  & -       
& ${10.06}$ & -       
& $0.123$  
\\
FlowNet2-ft~\cite{flownet2}
& $(1.45)$  & $4.16$  
& $(2.01)$  & $5.74$  
& $(1.28)$  & -  
& $(2.30)$  & $11.48\%$  
& $0.123$  
\\
SpyNet~\cite{spynet}
& $4.12$  & $6.69$ 
& $5.57$  & $8.43$ 
& $9.12$  & -  
& -       & -  
& $\mathbf{0.016}$  
\\
SpyNet-ft~\cite{spynet}
& $(3.17)$ & $6.64$ 
& $(4.32)$ & $8.36$  
& $(4.13)$ & $12.31\%$  
& -        &  $35.07\%$  
& $\mathbf{0.016}$  
\\
PWC-Net~\cite{pwcnet}
& $2.55$  & -  
& $3.93$  & -  
& $4.14$  & -  
& $10.35$ & $33.67\%$ 
& $0.030$  
\\
PWC-Net-ft~\cite{pwcnet}
& $(2.02)$  & $4.39$   
& $(2.08)$  & $\mathbf{5.04}$  
& -         & ${4.22\%}$  
& $(2.16)$  & $9.80\%$	  
& $0.030$  
\\


\headerline 
\textbf{With joint occ est.}
&&&&&&&&&\\
\headerline 
MirrorFlow~\cite{mirrorflow}
& -  & $3.32$  
& -  & $6.07$  
& -  & $4.38\%$  
& -  & $10.29\%$  
& $\pz\pz660$
\\
S2D flow~\cite{affine_s2d}
& -  & $18.48$  
& -  & $6.82$  
& -  & -  
& -  & -  
& $\pz2280$ 
\\

Our FlowNet-CSS
& $2.08$  & $3.94$  
& $3.61$  &  $6.03$  
& $3.69$  & -  
& $\mathbf{9.33}$  & -  
& $0.068$ 
\\
Our FlowNet-CSS-ft
& $(1.47)$  & $4.35$ 
& $(2.12)$  & $5.67$
& $(1.19)$  & $\mathbf{3.45}\%$  
& $(1.79)$  & $\mathbf{8.60}\%$  
& $0.068$ 
\\
Our FlowNet-css 
& $2.65$  & -  
& $4.05$  & -  
& $5.05$  & -  
& $11.74$ &    
& $0.033$  
\\

\hhline{|-||----|----||-|}
\end{tabular}
}

\vspace*{2mm}
\caption{
\label{tab:compare_flow}
Benchmark results for \textbf{optical flow estimation}. 
 We report the average endpoint error (AAE) for all benchmarks, except KITTI, where Out-noc and F1-all are used for the benchmark ranking on KITTI 2012 and 2015, respectively. Out-noc shows the percentage of outliers with errors more than $3$px in non-occluded regions, whereas F1-all shows the percentage in all regions. 
 Entries in parentheses denote methods that were finetuned on the evaluated dataset.
On the Sintel dataset, the performance of our networks is on par with FlowNet2. 
When comparing to other methods with joint occlusion estimation we are faster by multiple orders of magnitude.
On KITTI 2012 and 2015 we obtain state-of-the-art results among all optical flow methods (two frame, non-stereo)
}
\end{table}

\begin{table}[h!]
\centering


\begin{tabular}{|l| c | c | c | c| c|}
\hline
{\bf Method}                               &  {\bf D1-all} &  {\bf D2-all} & {\bf Fl-all} &  {\bf SF-all} & \bf Runtime (s) \\
\hline
ISF  \cite{isf}                         &   $4.46$      &   $5.95$      &   $6.22$        &  $8.08$        & $600$ \\
\hline
SGM+FlowFields (interp.)\cite{sgm_ff}      &   $13.37$     &   $27.80$     &   $22.82$    &  $33.57$     & $29$ \\
SceneFFields (dense)  \cite{scene_ff}      &   $6.57$      &   $10.69$     &   $12.88$    &  $15.78$     & $65$\\
Ours(interp.)                              &   $2.16$      &   $13.71$     &   $8.60$     &  $17.73$     & $0.22$ \\
Ours(dense)                                &   $2.16$      &   $6.45$      &   $8.60$     &  $11.34$     & $0.25$ \\
\hline 
\end{tabular}
\vspace*{4mm}
\caption{Benchmark results for \textbf{scene flow estimation}. "Interp." means the disparity values were automatically interpolated by the KITTI benchmark suite in the sparse regions. Compared to~\cite{scene_ff} we obtain much improved results and close the performance gap to much slower state-of-the-art methods, such as \cite{isf}, which use 2D information by a large margin
\label{tab:compare_scene} 
}
\end{table}

\section{Conclusion} 
We have shown that, in contrast to traditional methods, CNNs can very easily estimate occlusions and depth or motion boundaries, and that their performance surpasses traditional approaches by a large margin. 
While classical methods often use the backward flow to determine occlusions, we have shown that a simple extension from the forward FlowNet 2.0 stack performs best in the case of CNNs. 
We have also shown that this generic network architecture performs well on the tasks of disparity and flow estimation itself and yields state-of-the-art results on benchmarks. Finally, we have shown that the estimated occlusions can significantly improve motion segmentation. 

\section*{Acknowledgements} 
We acknowledge funding by the EU Horizon2020 project TrimBot2020 and by Gala Sports, and donation of a GPU server by Facebook.
Margret Keuper acknowledges funding by DFG grant KE 2264/1-1.

\bibliographystyle{splncs04}
\bibliography{egbib}

\clearpage

\thispagestyle{empty}
\vspace*{5pt}
\begin{center}
{\Large \bf{
Supplementary Material
}
}
\end{center}
\vspace*{10pt}

\setcounter{section}{0}
\setcounter{figure}{0}
\setcounter{table}{0}
\setcounter{equation}{0}

\section{Video}
Please see the supplementary video for qualitative results on a number of video sequences at \url{https://www.youtube.com/watch?v=SwOdSaBRysI}. 

\section{Visual Examples} 

Figures~\ref{fig:real_gallery1},~\ref{fig:real_gallery2} and~\ref{fig:real_gallery3} give 
 examples of our motion boundary, occlusion and flow estimation on some real images. 
We show our FlowNet-CSS and our FlowNet-CSSR-ft-sd, which includes fine-tuning on small displacements 
and the final refinement network. One can observe that already FlowNet-CSS performs well, 
while FlowNet-CSSR-ft-sd provides smoother flow estimations and a bit more details. 
Generally, one can observe that motion boundaries are estimated well, indicating the usefulness 
for motion segmentation. This is also visible in the provided video. 

\begin{figure} 
    \begin{center}
    \resizebox{\linewidth}{!}{%
    \begin{tabular}{cccc}%
        \labelimage{0.25}{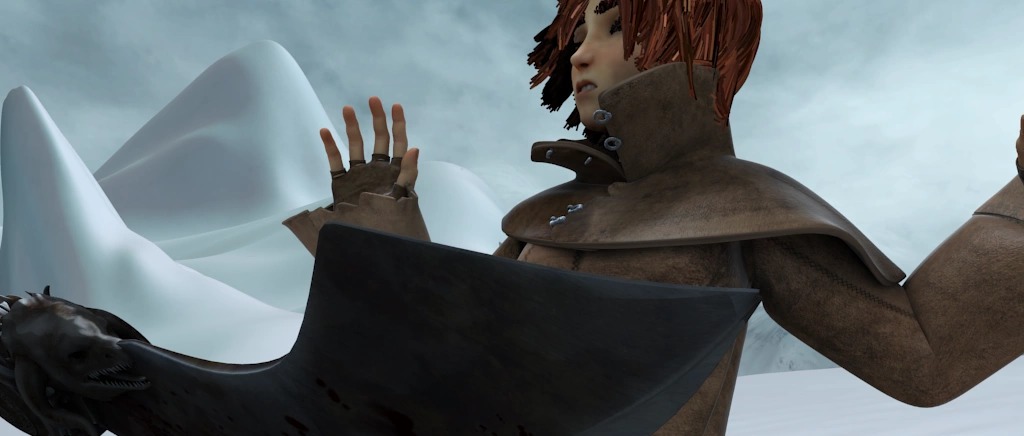}{Image L}        
        & \labelimage{0.25}{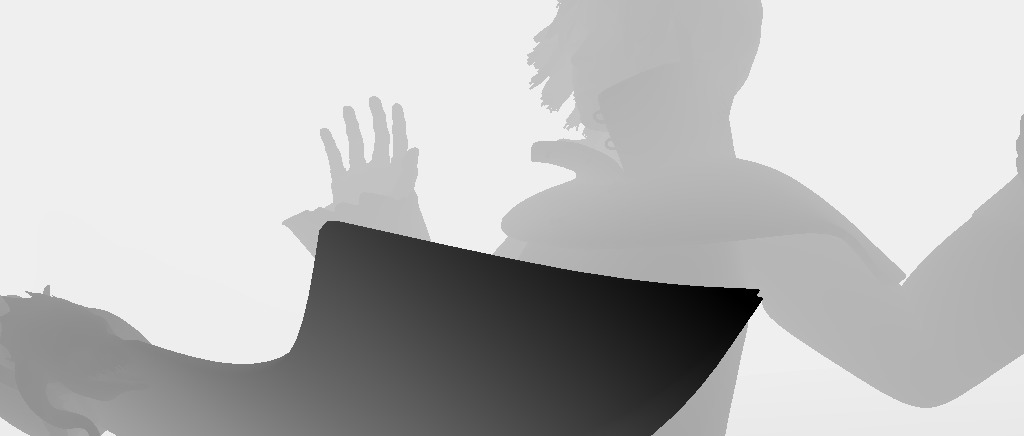}{GT Disp}  
        & \labelimage{0.25}{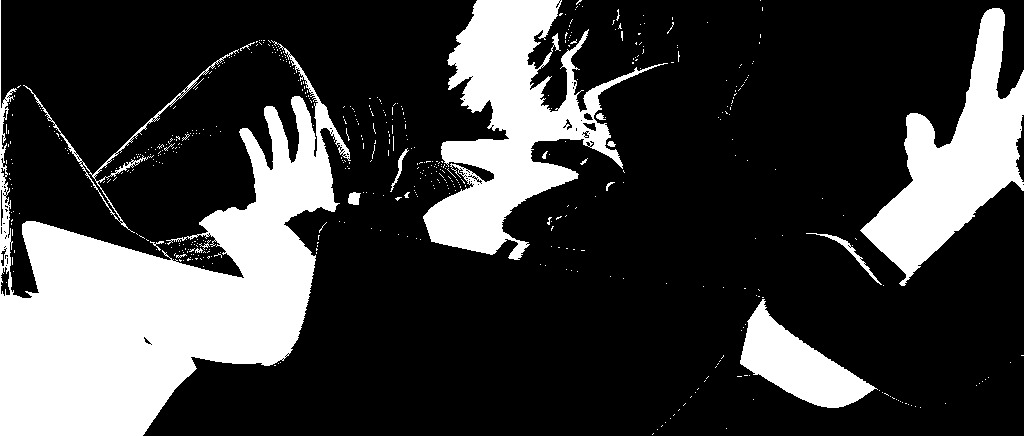}{GT Occ.}  
        & \labelimage{0.25}{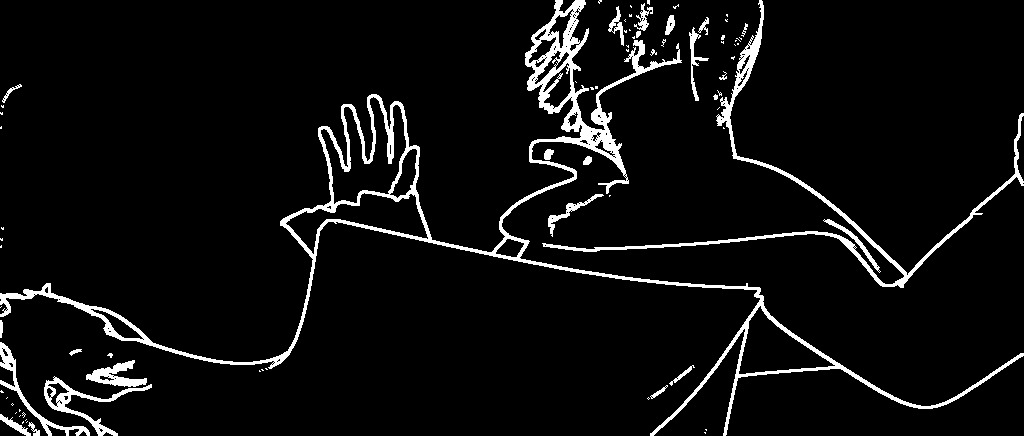}{GT D.Bnd.}  
        \\
        \labelimage{0.25}{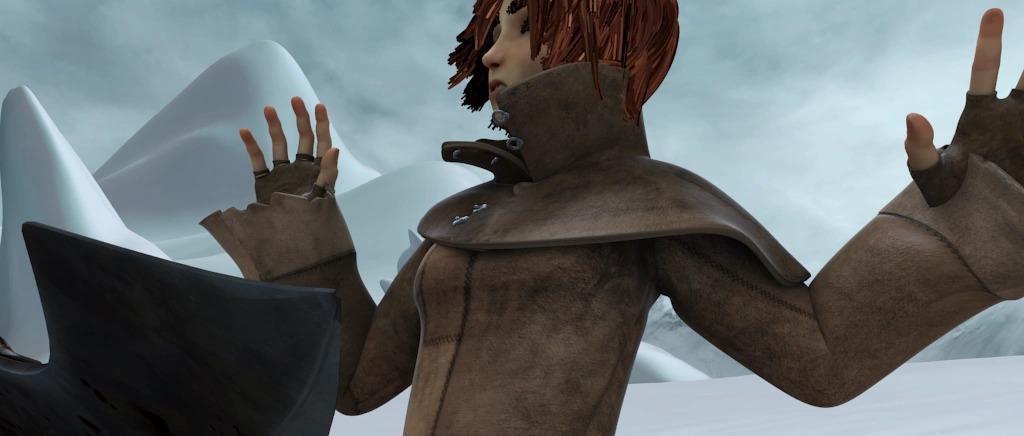}{Image R}        
        & \labelimage{0.25}{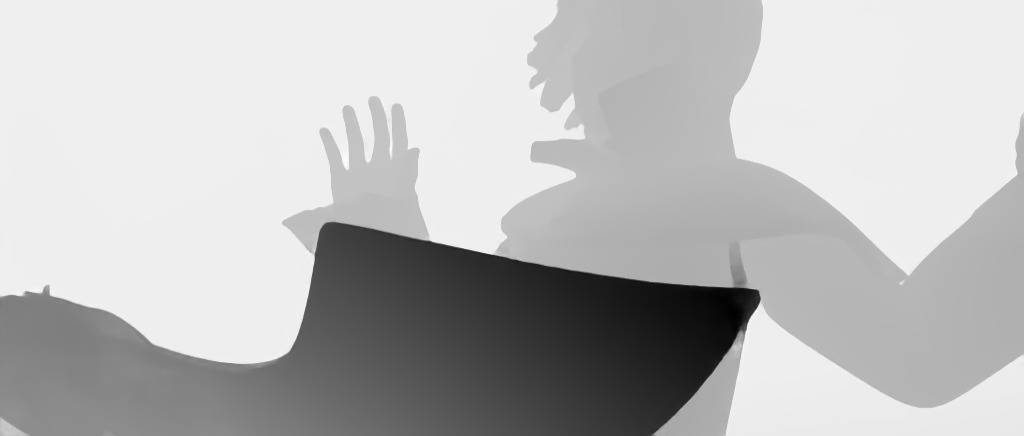}{Disp}  
        & \labelimage{0.25}{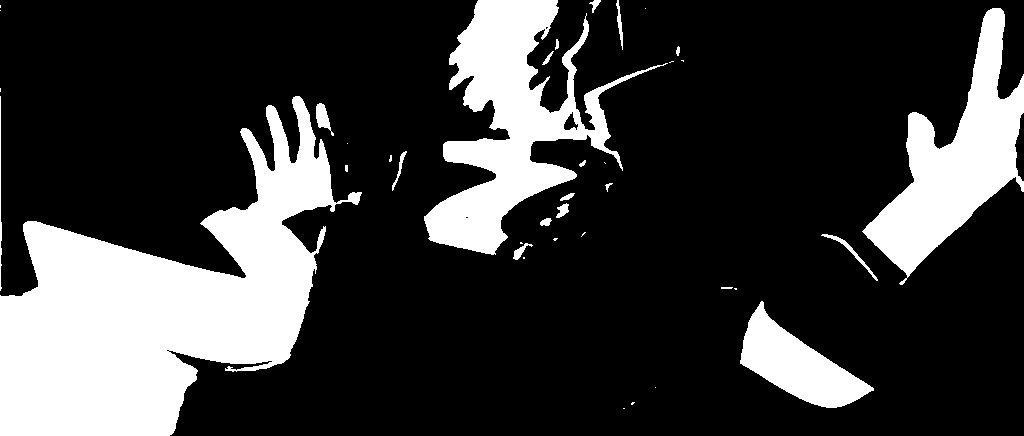}{Occ.}  
        & \labelimage{0.25}{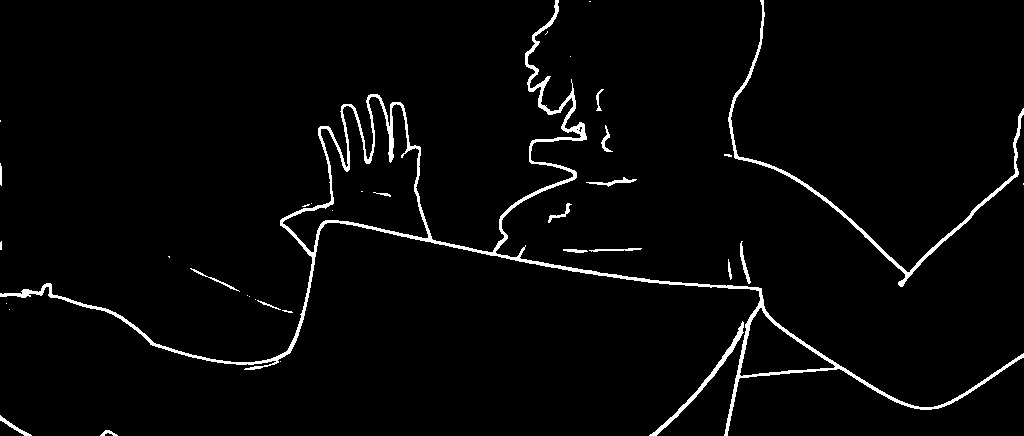}{D.Bnd.}  
        \\
        \hline 
        \vspace*{-2.5mm}%
        \\ 
        \labelimage{0.25}{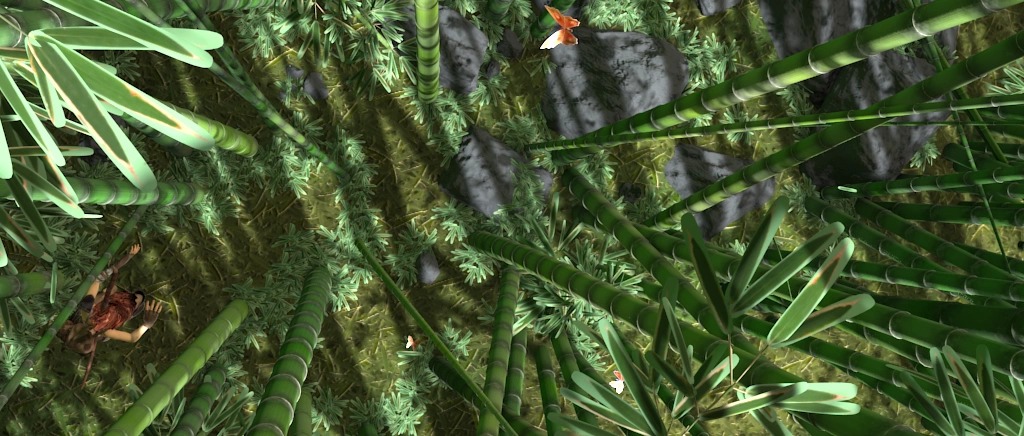}{Image L}        
        & \labelimage{0.25}{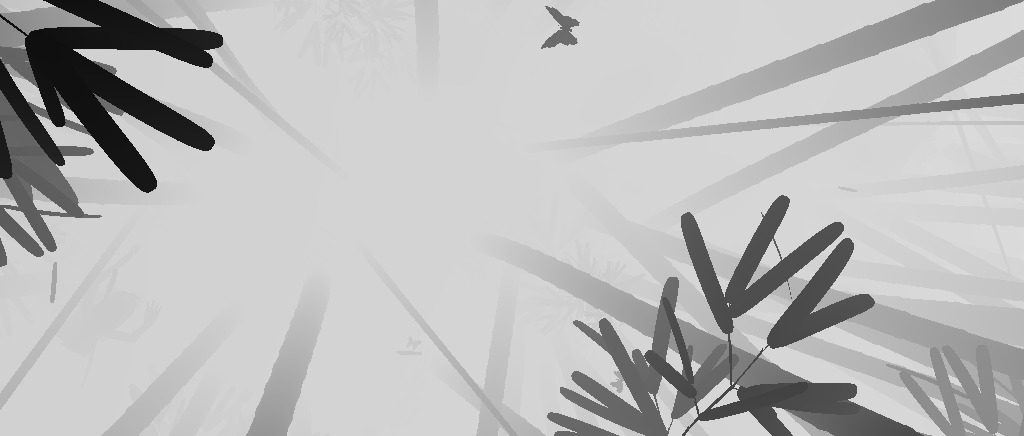}{GT Disp}  
        & \labelimage{0.25}{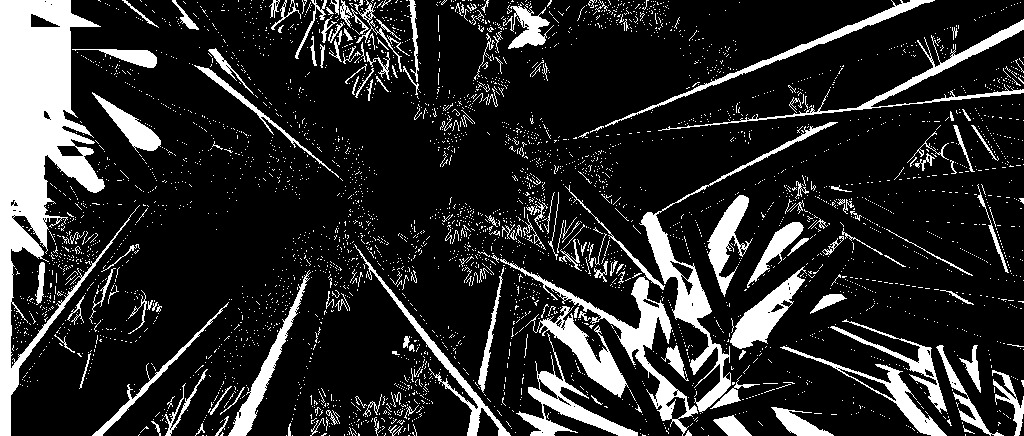}{GT Occ.}  
        & \labelimage{0.25}{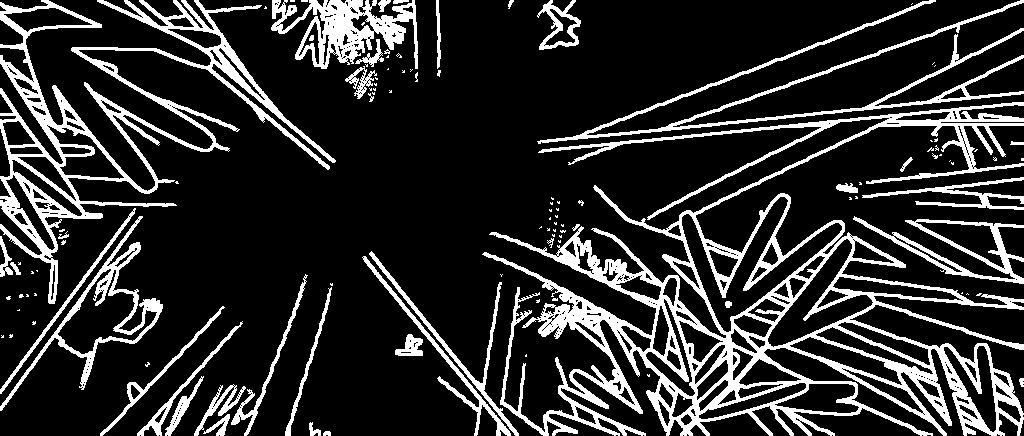}{GT D.Bnd.}  
        \\
        \labelimage{0.25}{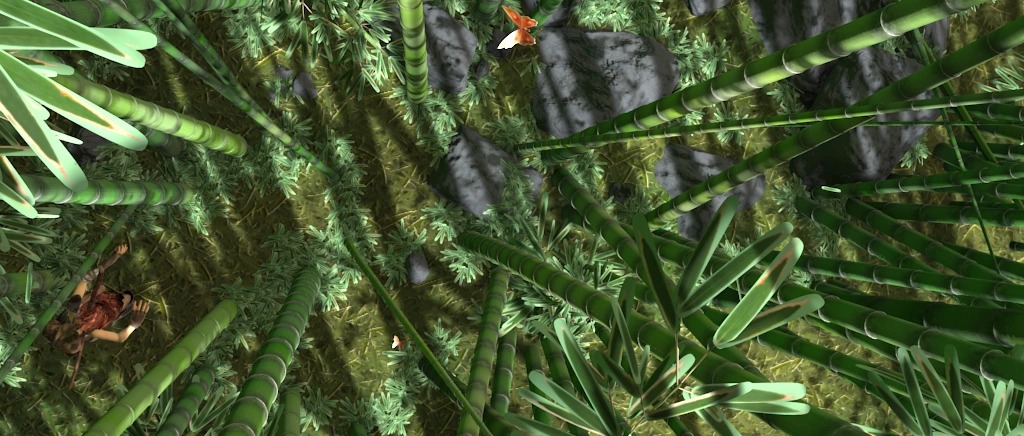}{Image R}        
        & \labelimage{0.25}{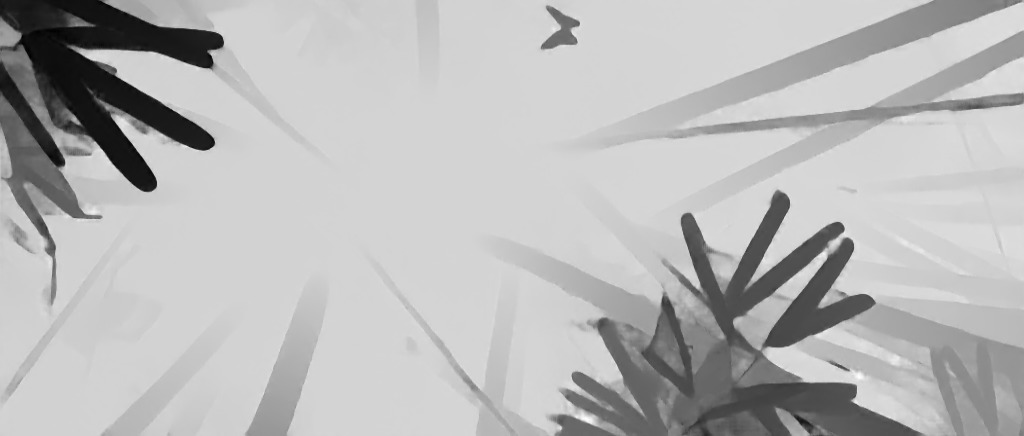}{Disp}  
        & \labelimage{0.25}{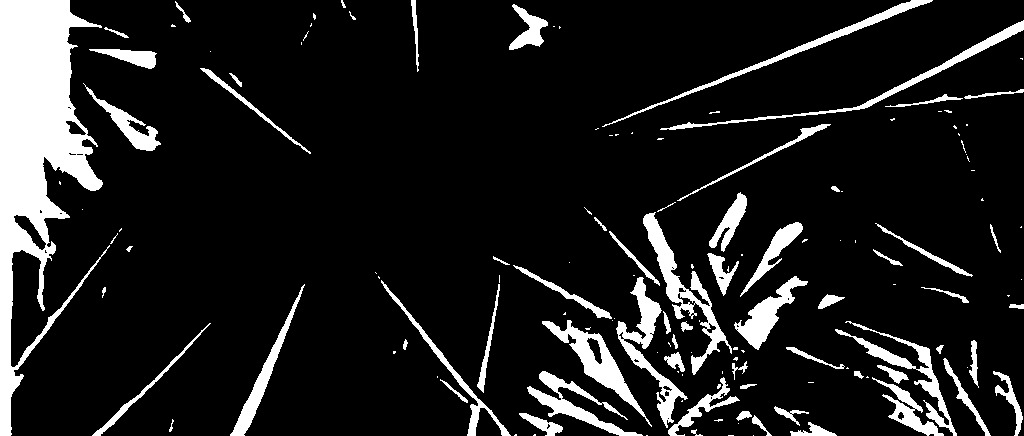}{Occ.}  
        & \labelimage{0.25}{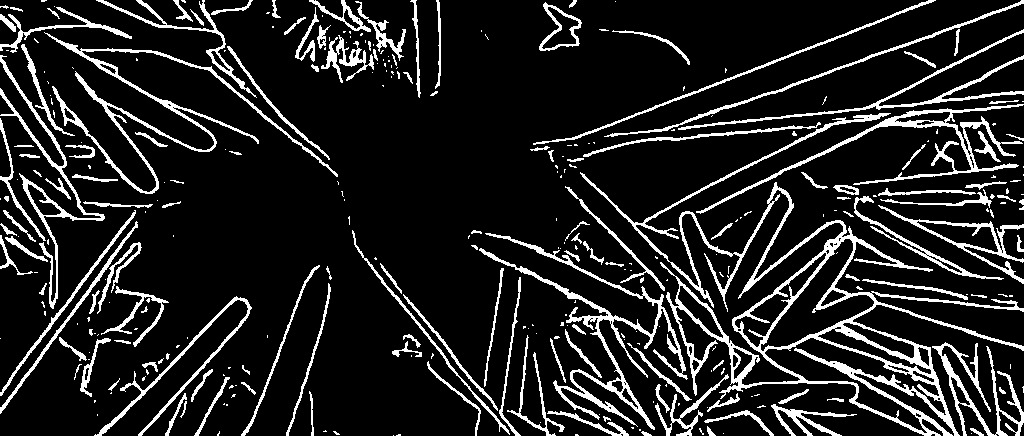}{D.Bnd.}  
        \\
    \end{tabular}  
    }
    \end{center} 
  \caption{ 
    Examples from our DispNet-CSS for joint depth boundary, occlusion and disparity estimation
    on some Sintel images. The estimations from our method are in general 
    very close to the ground-truth
    \label{fig:disp_gallery1}
  }
\end{figure}

Figure~\ref{fig:disp_gallery1} shows the case for depth boundary, occlusion and disparity estimation 
on some examples from Sintel. One can observe that the estimations are very close to the ground-truth 
and hard to distinguish at first glance. Note how well the DispNet-CSS architecture can estimate the 
large occlusion areas in the first, and the fine details in the second example. 

\begin{figure} 
    \begin{center}
    \begin{tabular}{ccc}%
         Ours (Interp.)
         & Ours (Dense) 
         & ISF \cite{isf}
         \\
         \labelimage{0.31}{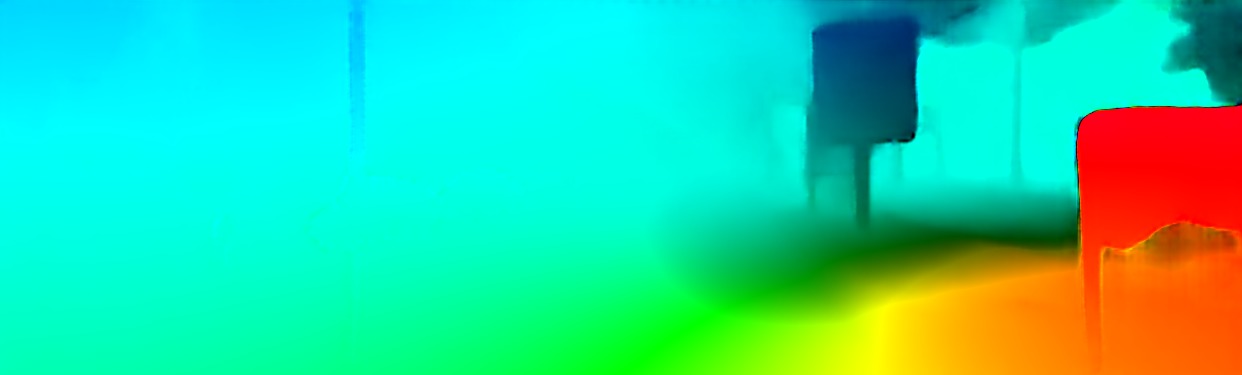}{Flow}  
        & \labelimage{0.31}{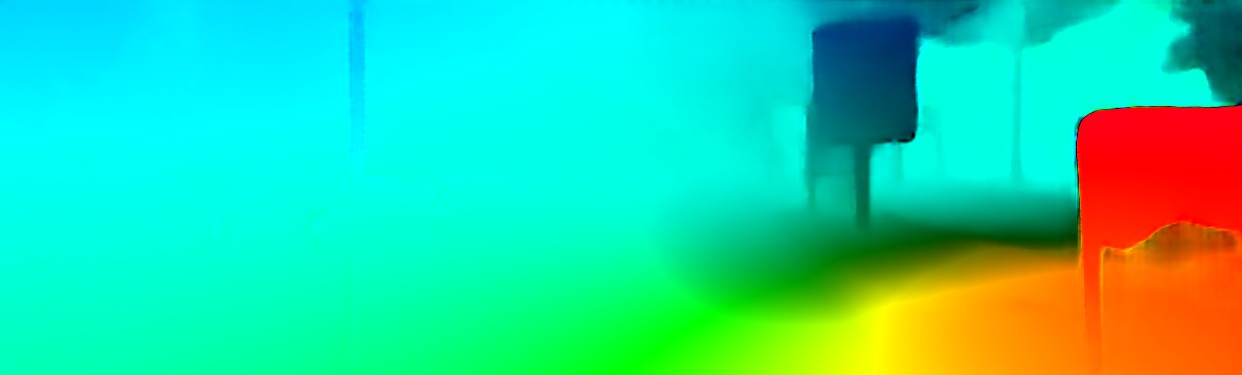}{Flow} 
        & \labelimage{0.31}{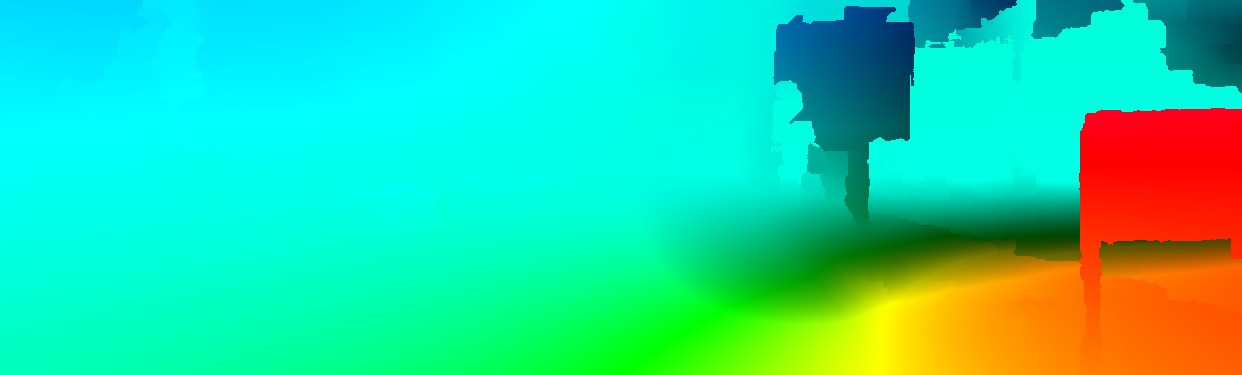}{Flow}
        \\
         \labelimage{0.31}{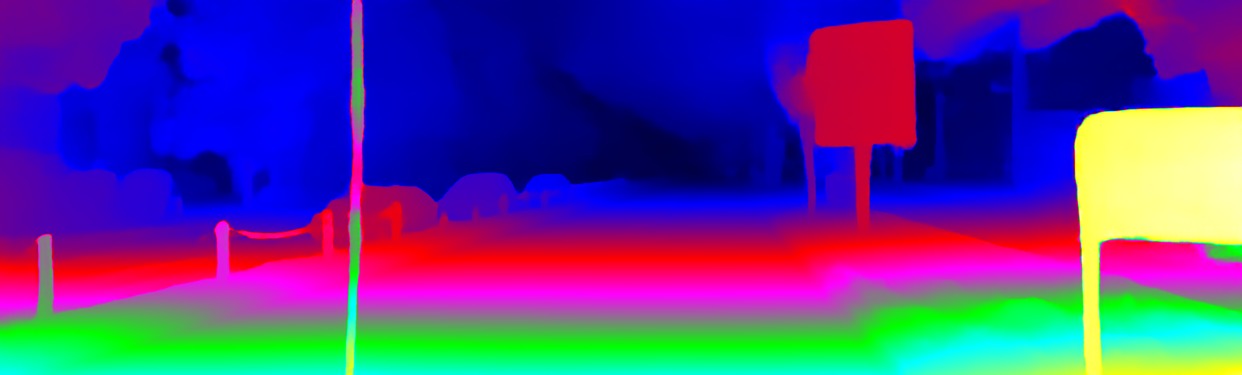}{Disp t=0}  
        & \labelimage{0.31}{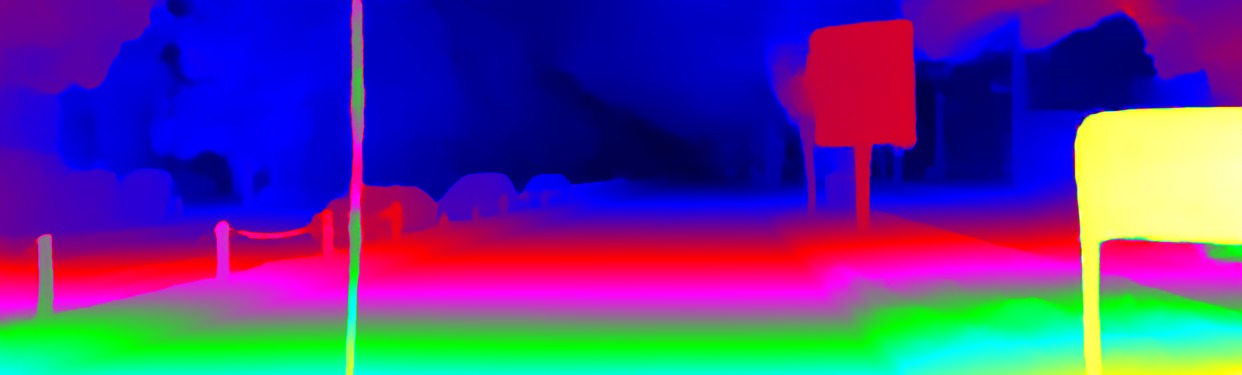}{Disp t=0} 
        & \labelimage{0.31}{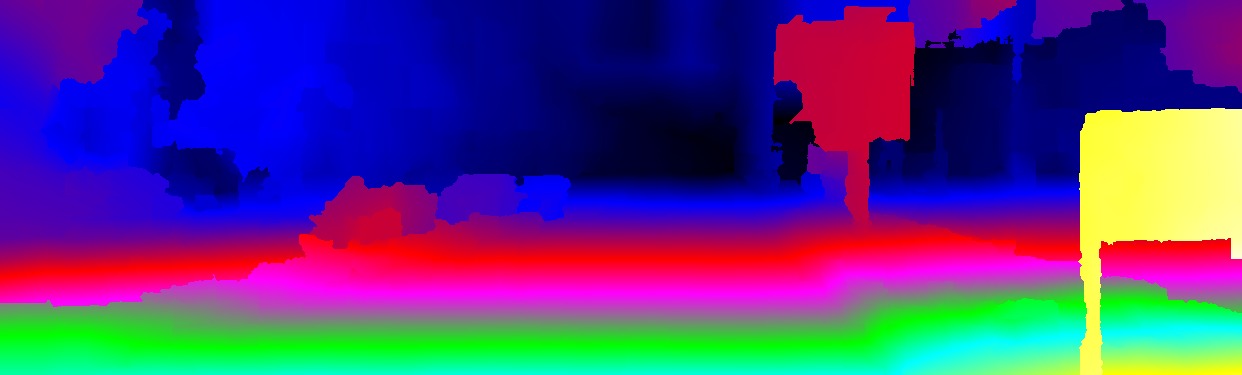}{Disp t=0}  
        \\
         \labelimage{0.31}{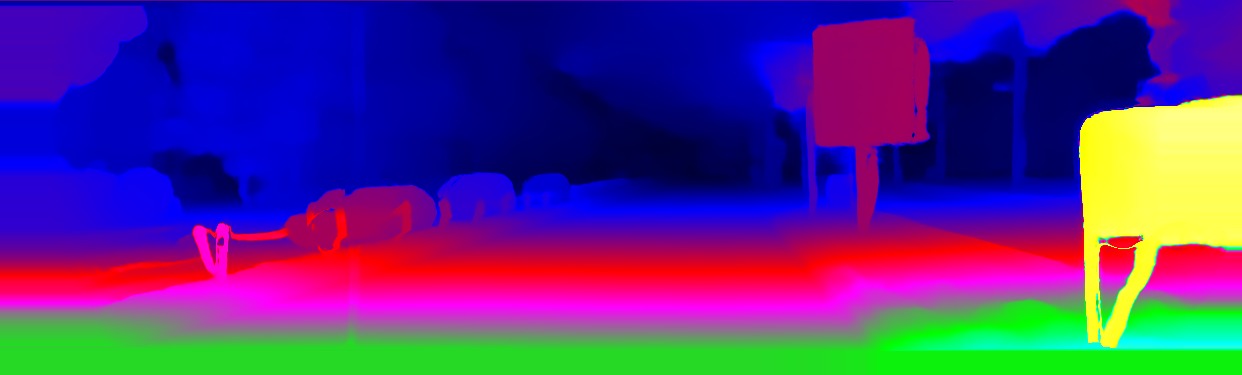}{Disp warped from t=1}  
        & \labelimage{0.31}{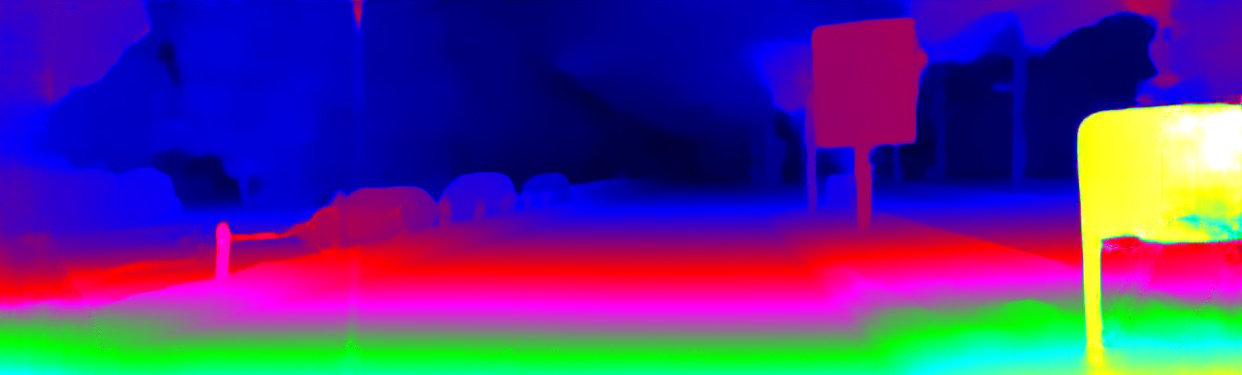}{Disp warped from t=1} 
        & \labelimage{0.31}{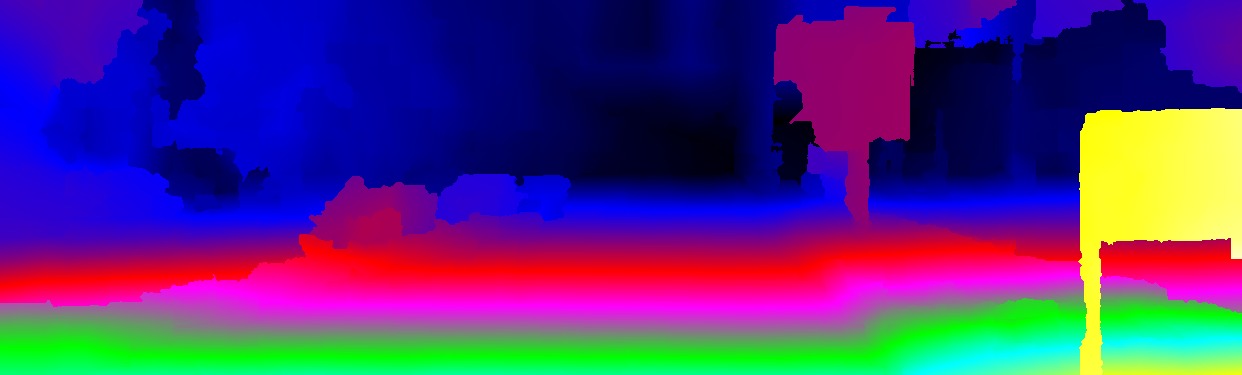}{Disp warped from t=1}
        \\ 
         \labelimage{0.31}{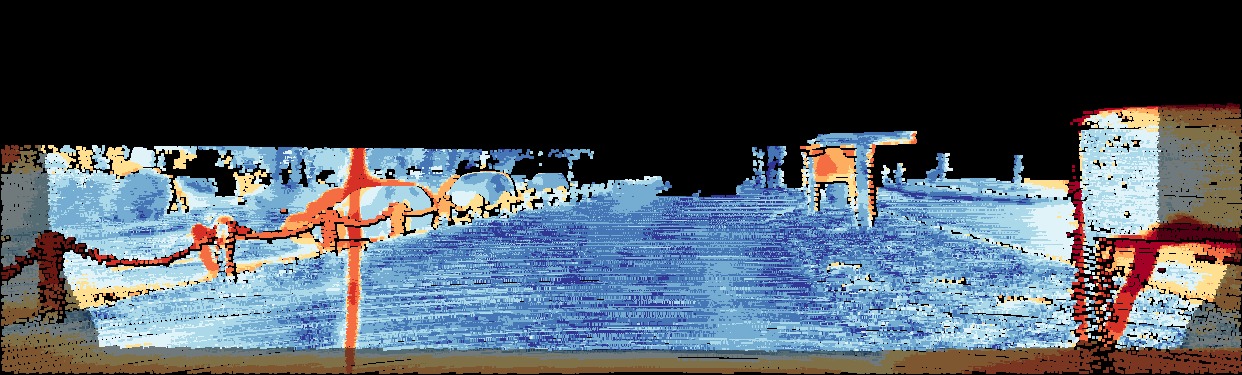}{SF error}  
        & \labelimage{0.31}{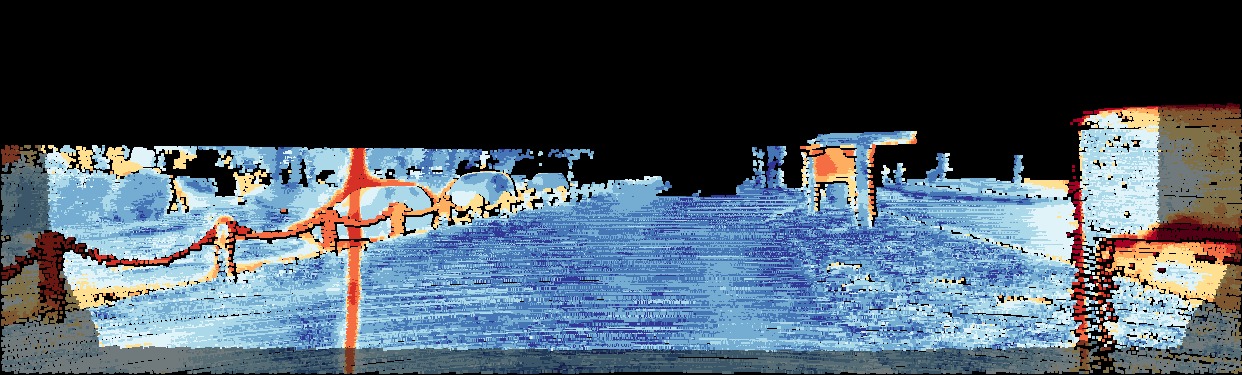}{SF error} 
        & \labelimage{0.31}{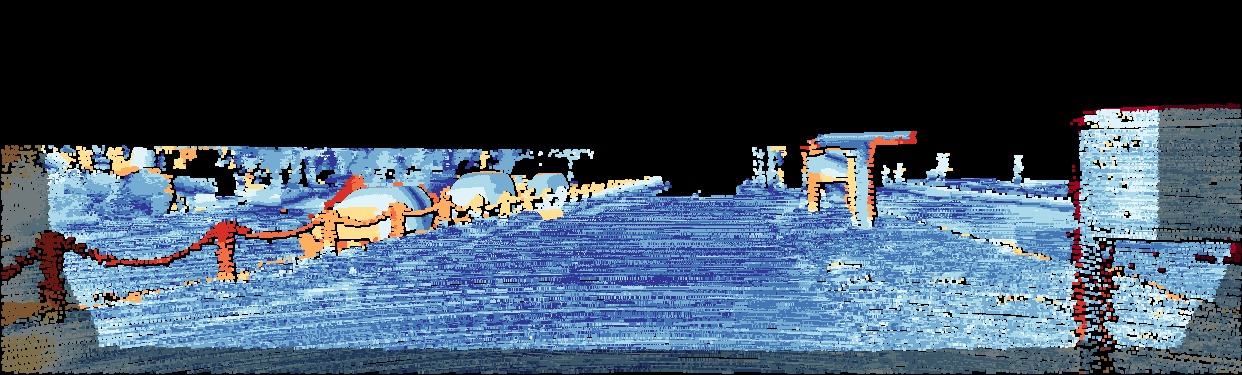}{SF error} 
    \end{tabular}  
    \end{center} 
   \caption{ 
     Example of our occlusion filling network for scene flow. We compare our results on scene flow estimation with the current state of art on KITTI~\cite{isf} (right) directly with the visualizations from the KITTI benchmark. 
     The first row shows the estimated optical flow on the left images from $t=0$ to $t=1$. 
     The second row shows disparity at $t=0$ with left image as the reference frame.
     The third row shows the disparity at $t=1$ warped to $t=0$ using in the forward flow (shown in the first row). 
     The last row shows the scene flow error map from the KITTI benchmark, where the occluded regions have a dark overlay.\newline 
     The first column shows results of sparse predictions. The interpolation into occluded regions is the default from the KITTI benchmark. 
     The second column shows the results when using our additional network to fill the occlusion areas and the third column visualizes the results from ISF~\cite{isf}. 
     We can observe that our scene flow architecture learns to fill reasonable disparity values in the occluded regions. The error in the bottom occlusion area is significantly lower. 
     Also, note that the hallucination effects at the road sign (yellow) are removed by our network
     \label{fig:sf_gallery1}
   }
\end{figure}

Figure~\ref{fig:sf_gallery1} shows results from the network proposed for filling occluded areas for 
scene flow. Note that occlusion in general causes hallucination effects (visible e.g. for the trees on the right)
and missing values at the boundaries. One can observe that our network is able to remove the hallucination effects 
and that it provides a meaningful extrapolation to the missing values. 

\section{Losses}

\subsection{Scaling of Predictions \label{sec:loss_scaling}} 

\begin{figure*}[t]
    \begin{center}
        \subfigure[FlowNet2-C~\cite{flownet2}\label{fig:flownet_old}]{
            \includegraphics[width=0.3\textwidth]{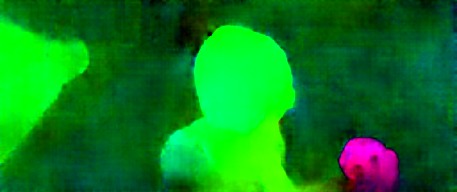}
        }
        \subfigure[Scaling\label{fig:flownet_rebuild}]{
            \includegraphics[width=0.3\textwidth]{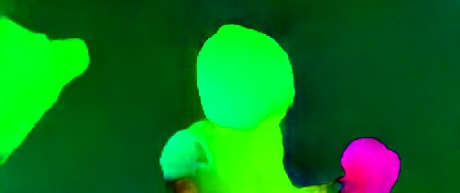}
        }
        \subfigure[Ground-truth]{
            \includegraphics[width=0.3\textwidth]{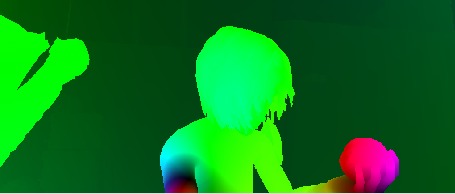}
        }
    \end{center}
    
    \caption{Comparison of FlowNets trained with different scalings on a Sintel example.
    (a) shows the first network of the stack published in FlowNet2~\cite{flownet2}.
    (b) shows the same network with the scaling as proposed by us. (c) shows the ground-truth. 
    Note how the scaling significantly removes noise for small displacements
    \label{fig:fn_samples}
    }
\end{figure*}

\begin{figure} 
    \begin{center}
    \begin{tabular}{ccc}%
        \labelimage{0.25}{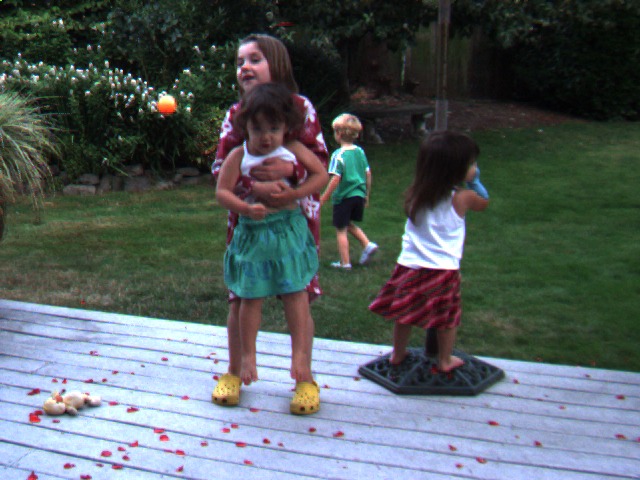}{Image 0} &     
        \labelimage{0.25}{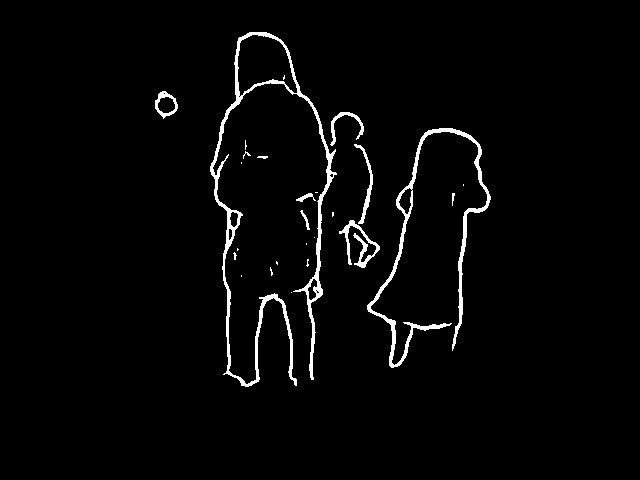}{M.Bnd.} &     
        \labelimage{0.25}{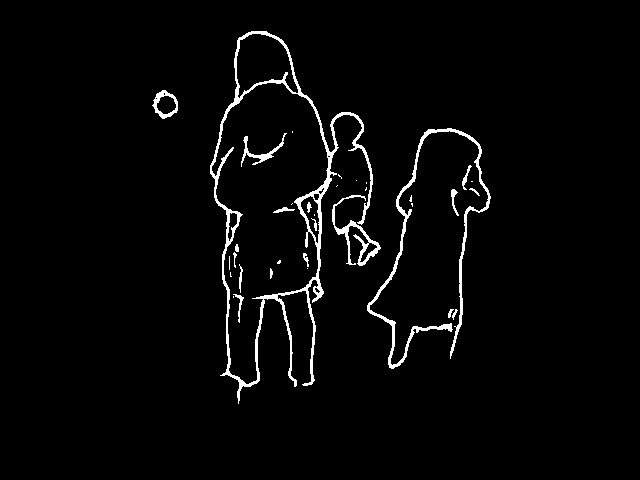}{M.Bnd. (ft-sd)} \\
        \labelimage{0.25}{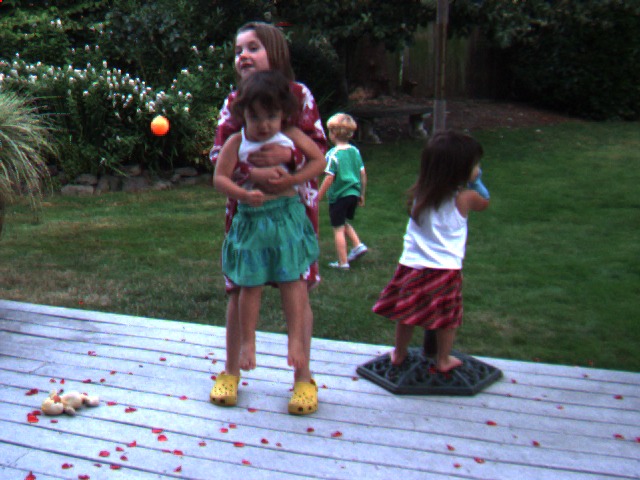}{Image 1} &     
        \labelimage{0.25}{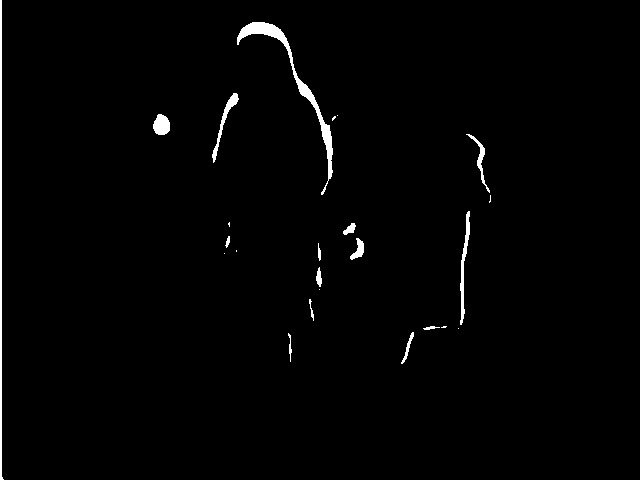}{Occ.} &     
        \labelimage{0.25}{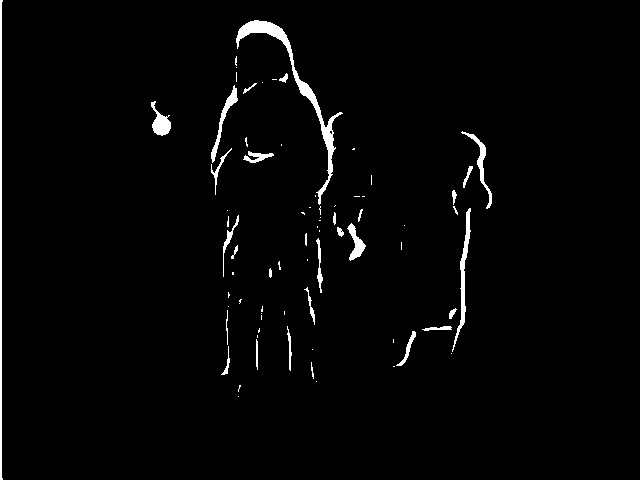}{Occ. (ft-sd)} \\
        \labelimage{0.25}{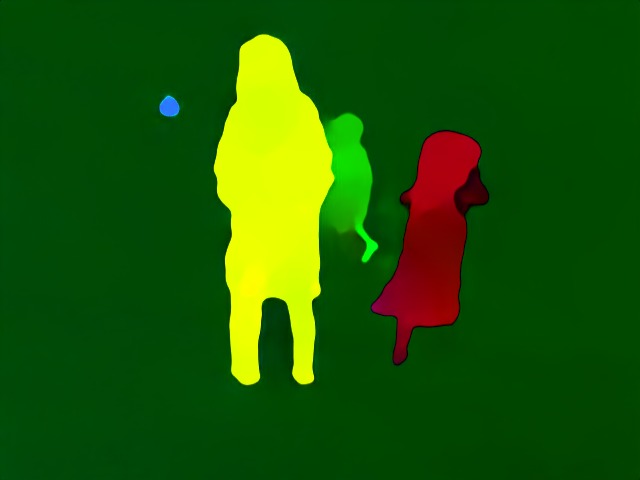}{FlowNet2} &     
        \labelimage{0.25}{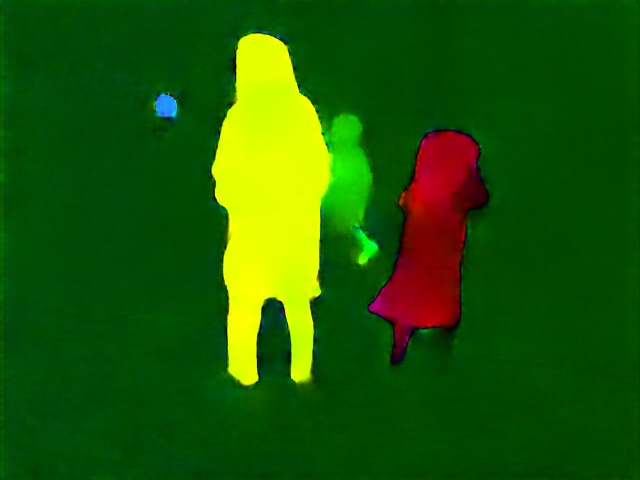}{Flow Ours} &     
        \labelimage{0.25}{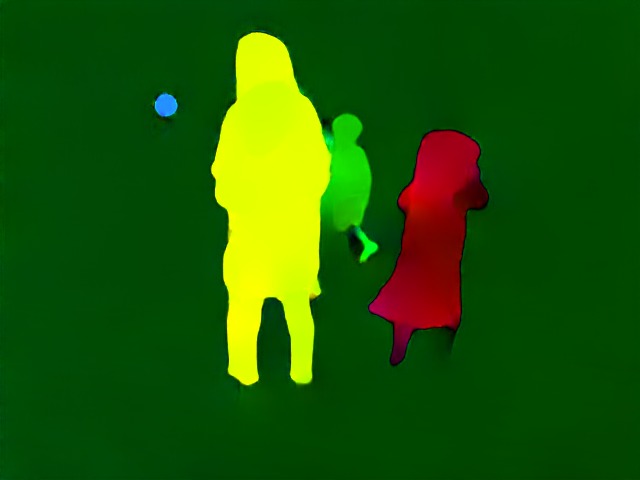}{Flow Ours (ft-sd)} \\
        \hline 
        \vspace*{-2.5mm}%
        \\ 
        \labelimage{0.25}{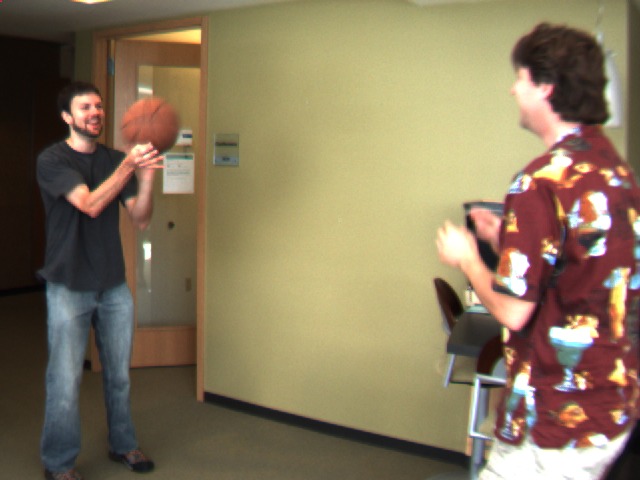}{Image 0} &     
        \labelimage{0.25}{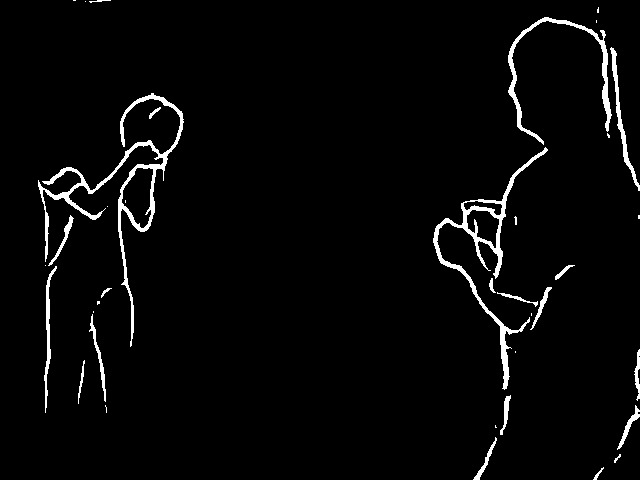}{M.Bnd.} &     
        \labelimage{0.25}{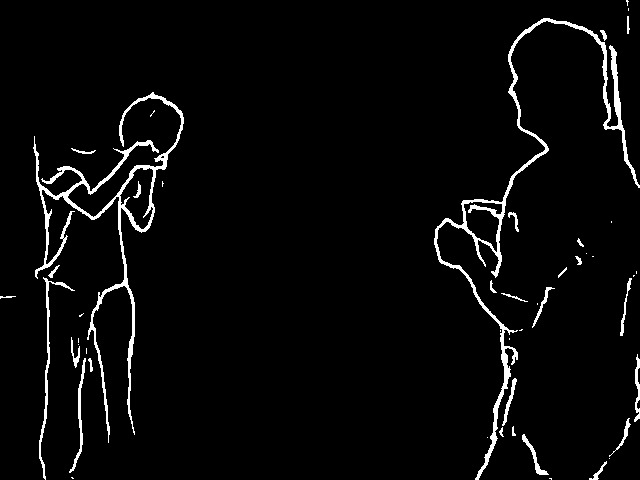}{M.Bnd. (ft-sd)} \\
        \labelimage{0.25}{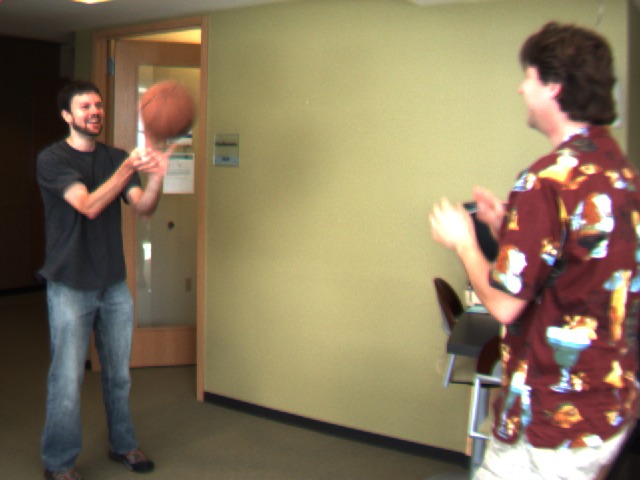}{Image 1} &     
        \labelimage{0.25}{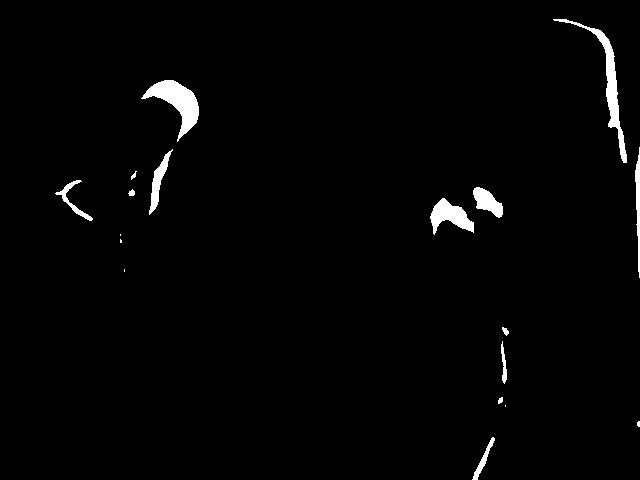}{Occ.} &     
        \labelimage{0.25}{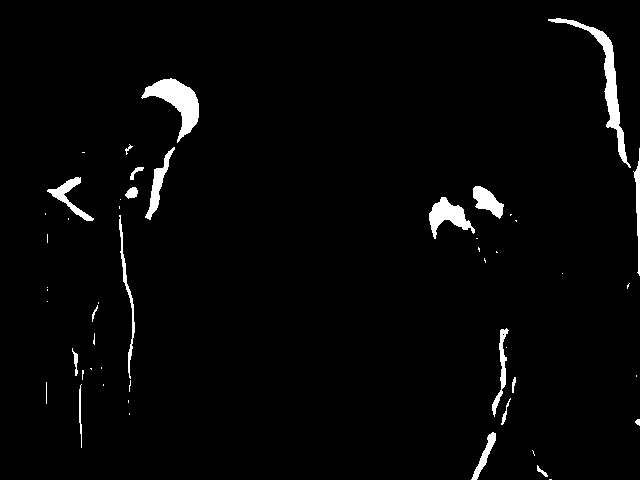}{Occ. (ft-sd)} \\
        \labelimage{0.25}{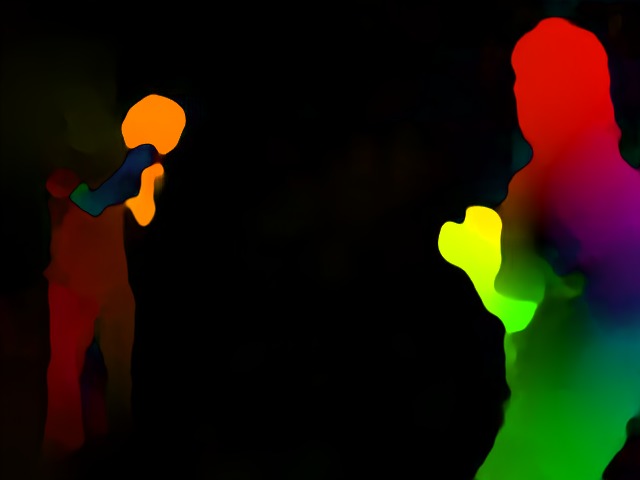}{FlowNet2} &     
        \labelimage{0.25}{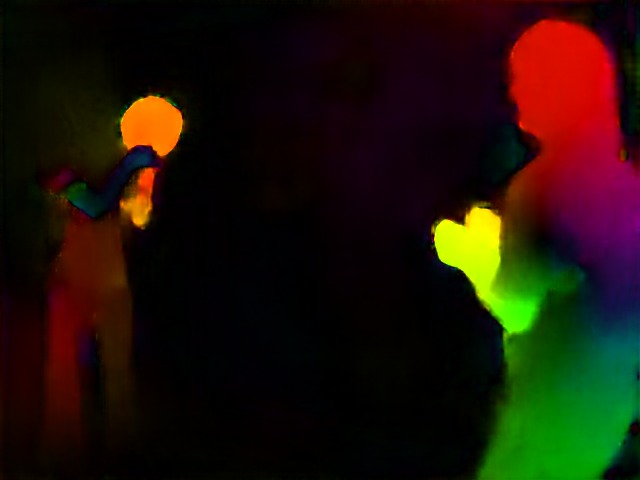}{Flow Ours} &     
        \labelimage{0.25}{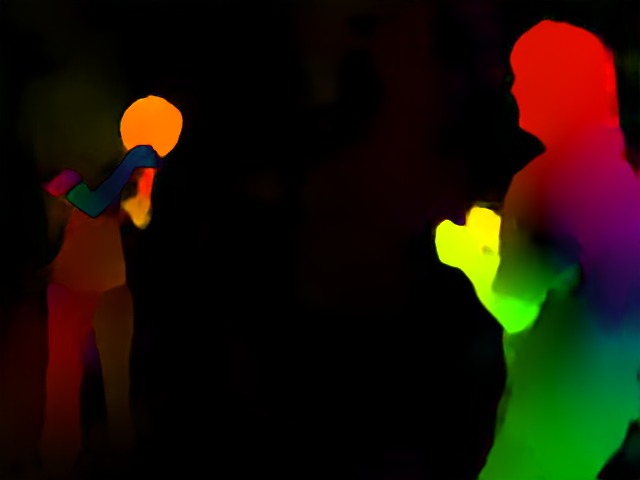}{Flow Ours (ft-sd)} \\
    \end{tabular}  
    \end{center} 
  \caption{
    Examples of our joint motion boundary, occlusion and optical flow estimation
    on some real images. We provide estimations for FlowNet-CSS and FlowNet-CSSR-ft-sd. 
    One can observe that our method provides very sharp estimations and that the 
    version fine-tuned for small displacements provides a bit more details. 
    In the top example the occlusions from 
    the child in the background become visible  
    \label{fig:real_gallery1}
  }
\end{figure}

\begin{figure} 
    \begin{center}
    \begin{tabular}{ccc}%
        \labelimage{0.25}{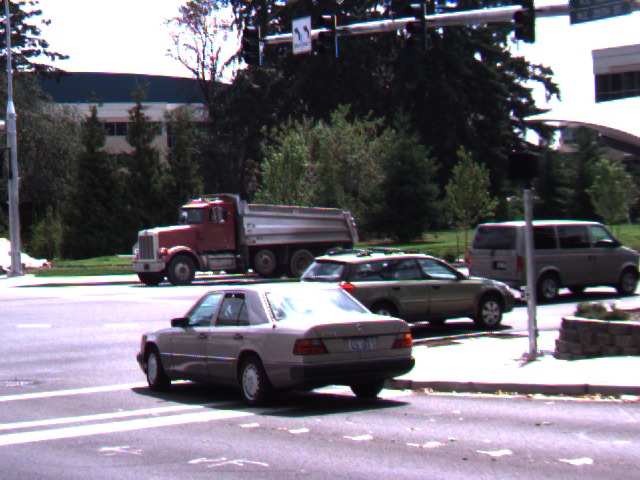}{Image 0} &     
        \labelimage{0.25}{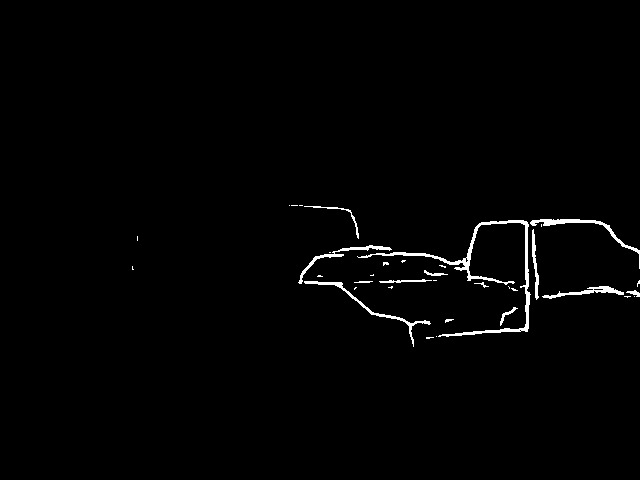}{M.Bnd.} &     
        \labelimage{0.25}{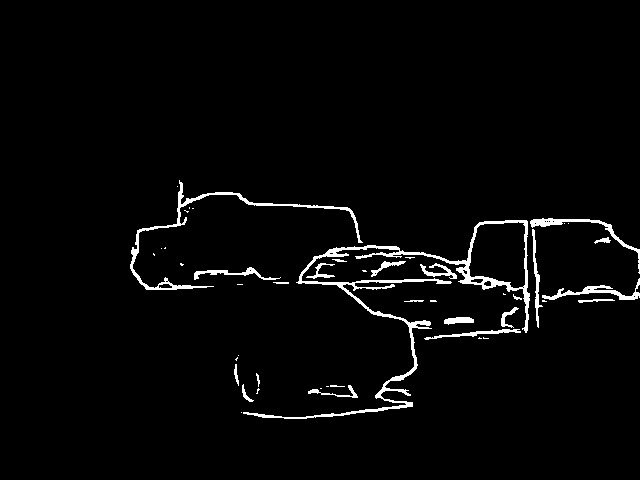}{M.Bnd. (ft-sd)} \\
        \labelimage{0.25}{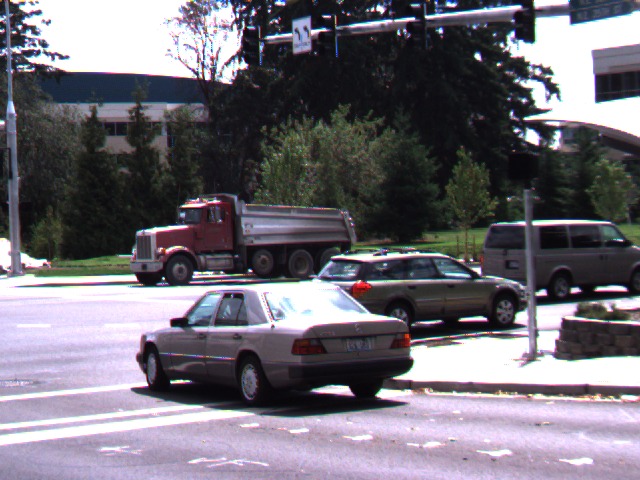}{Image 1} &     
        \labelimage{0.25}{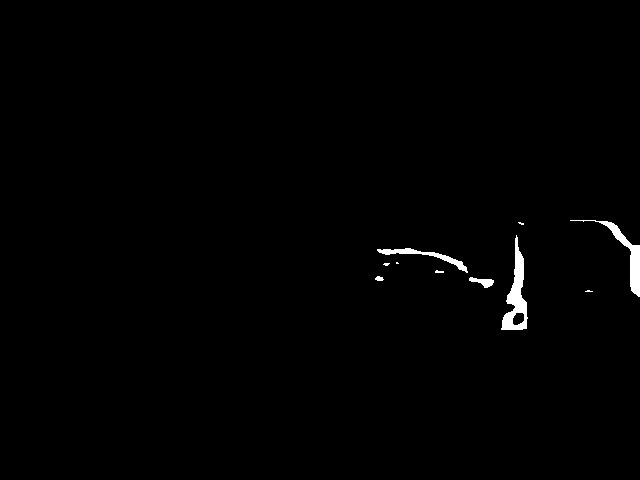}{Occ.} &     
        \labelimage{0.25}{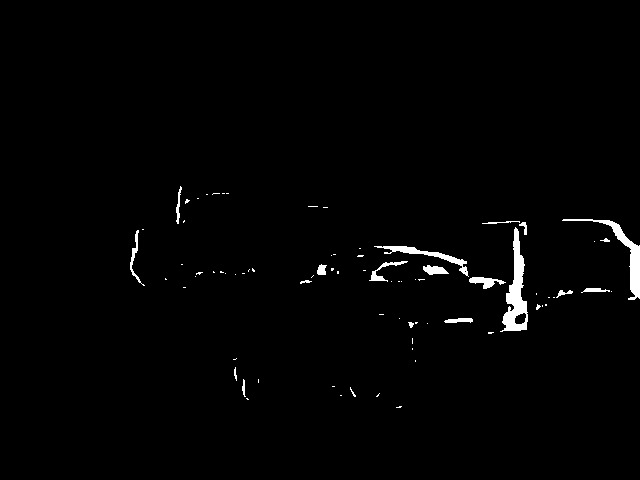}{Occ. (ft-sd)} \\
        \labelimage{0.25}{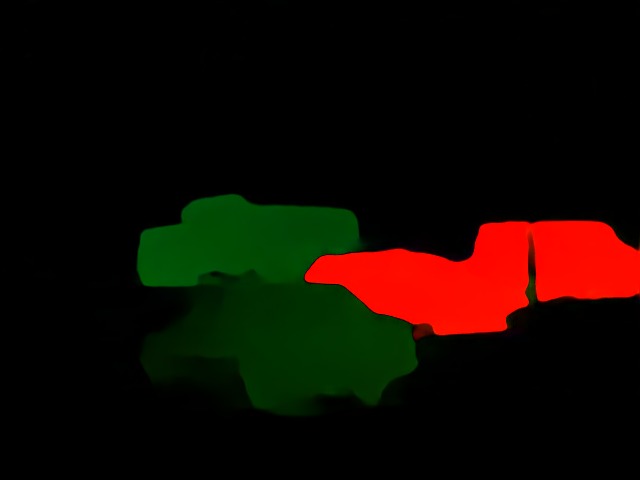}{FlowNet2} &     
        \labelimage{0.25}{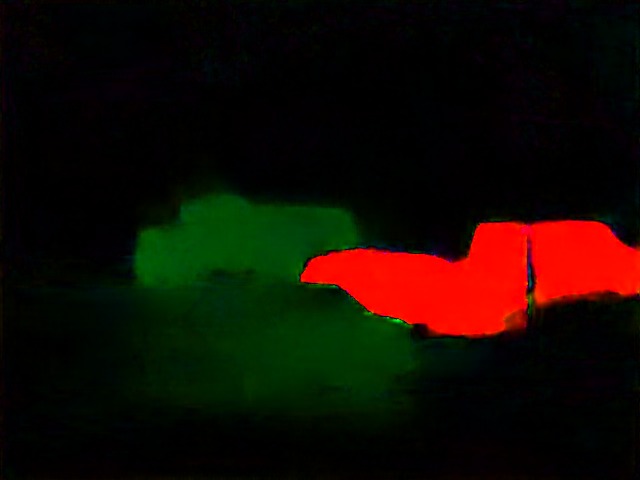}{Flow Ours} &     
        \labelimage{0.25}{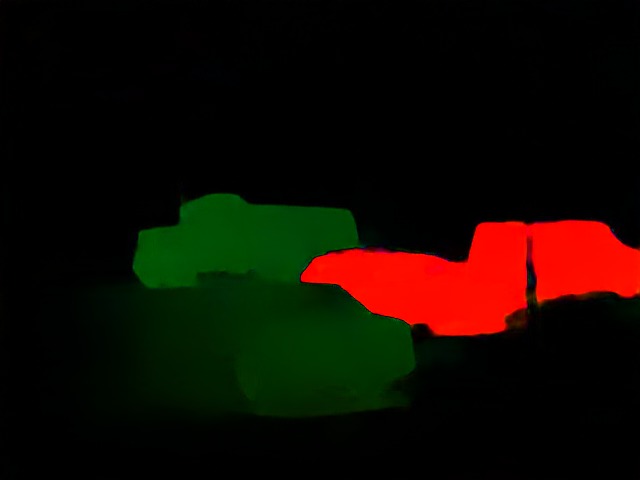}{Flow Ours (ft-sd)} \\
        \hline 
        \vspace*{-2.5mm}%
        \\ 
        \labelimage{0.25}{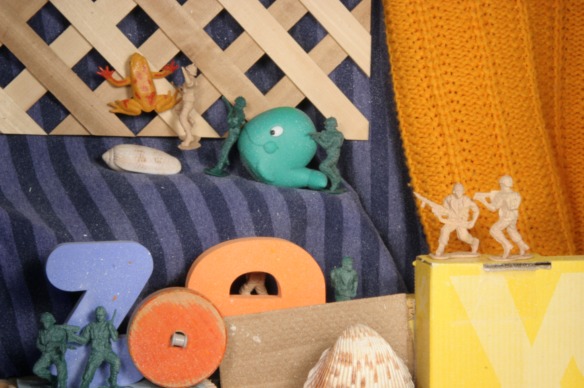}{Image 0} &     
        \labelimage{0.25}{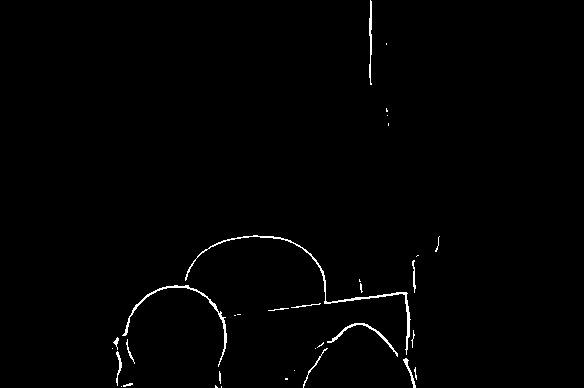}{M.Bnd.} &     
        \labelimage{0.25}{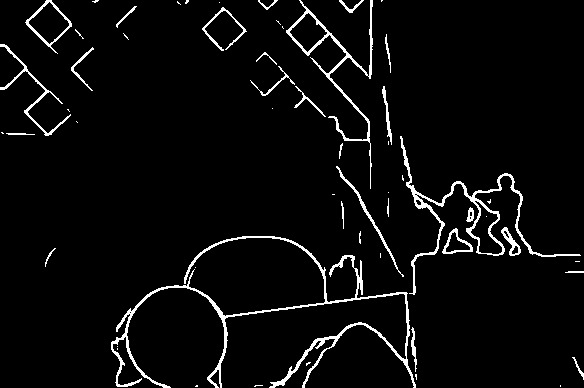}{M.Bnd. (ft-sd)} \\
        \labelimage{0.25}{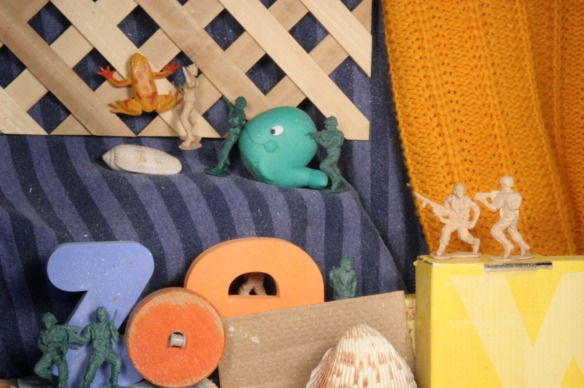}{Image 1} &     
        \labelimage{0.25}{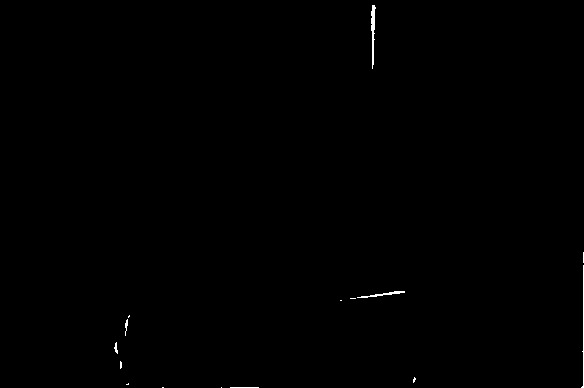}{Occ.} &     
        \labelimage{0.25}{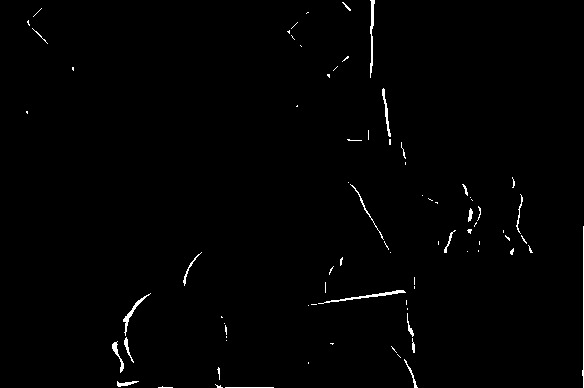}{Occ. (ft-sd)} \\
        \labelimage{0.25}{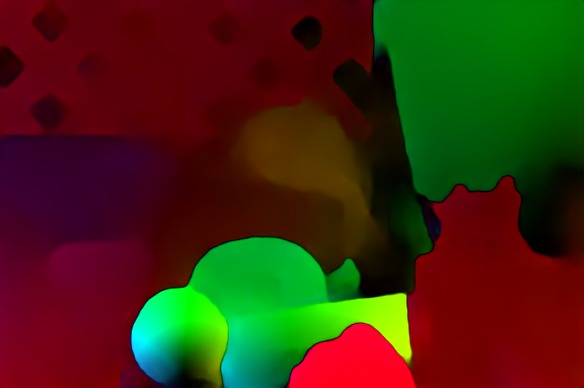}{FlowNet2} &     
        \labelimage{0.25}{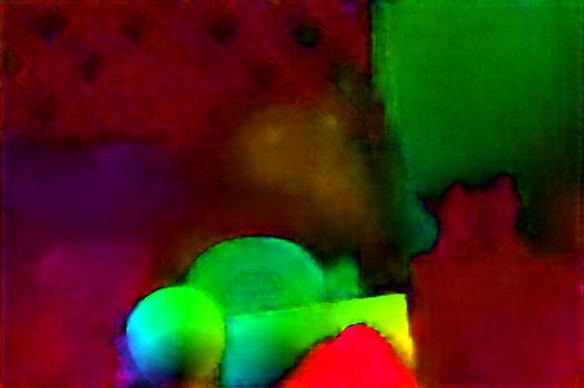}{Flow Ours} &     
        \labelimage{0.25}{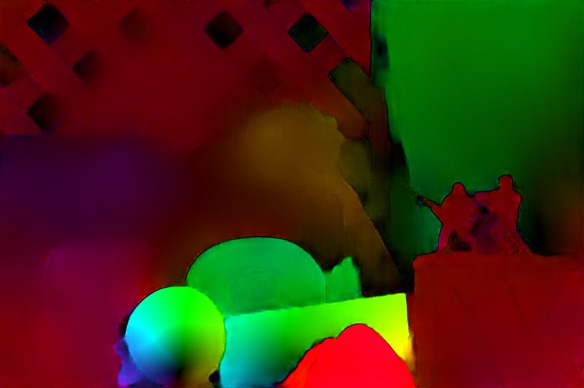}{Flow Ours (ft-sd)} \\
    \end{tabular}  
    \end{center} 
  \caption{
    More examples of our joint motion boundary, occlusion and optical flow estimation
    on some real images. We provide estimations for FlowNet-CSS and FlowNet-CSSR-ft-sd. 
    One can observe that our method provides very sharp estimations and that the 
    version fine-tuned for small displacements provides a bit more details. 
    In the top example, the boundaries of the cars become apparent. Note the fine details, 
    such as the exhaust pipe of the truck. In the bottom example, motion from the background pattern 
    is visible in the fine-tuned version 
    \label{fig:real_gallery2}
  }
\end{figure}

\begin{figure} 
    \begin{center}
    \begin{tabular}{ccc}%
        \labelimage{0.25}{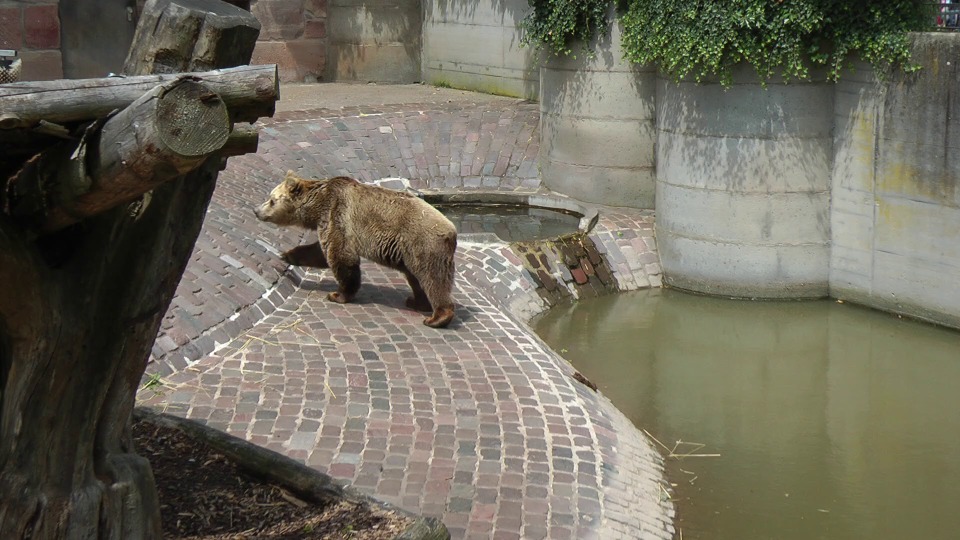}{Image 0} &     
        \labelimage{0.25}{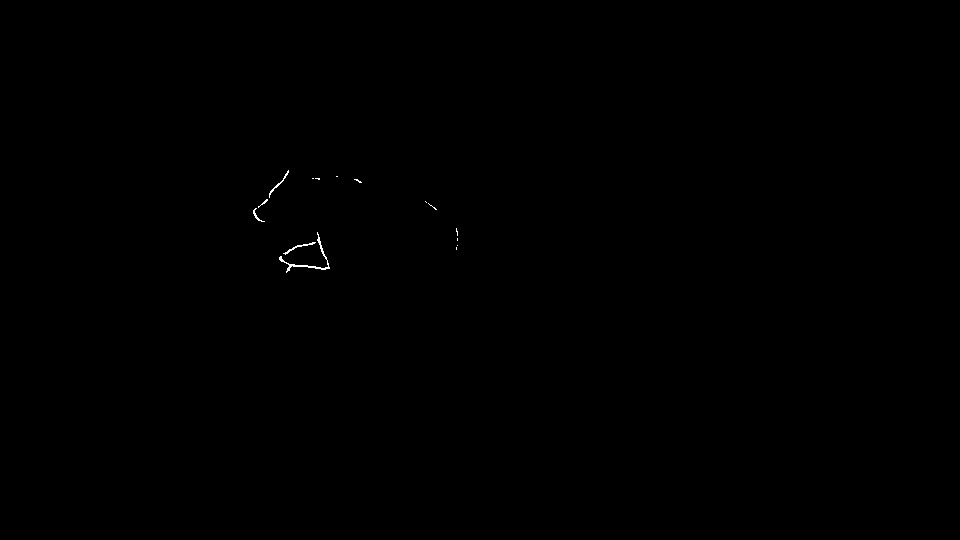}{M.Bnd.} &     
        \labelimage{0.25}{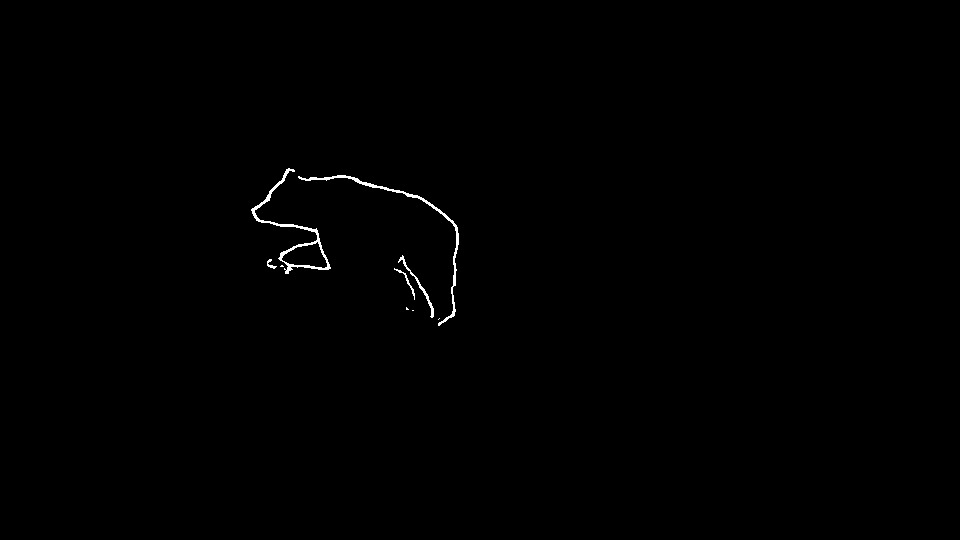}{M.Bnd. (ft-sd)} \\
        \labelimage{0.25}{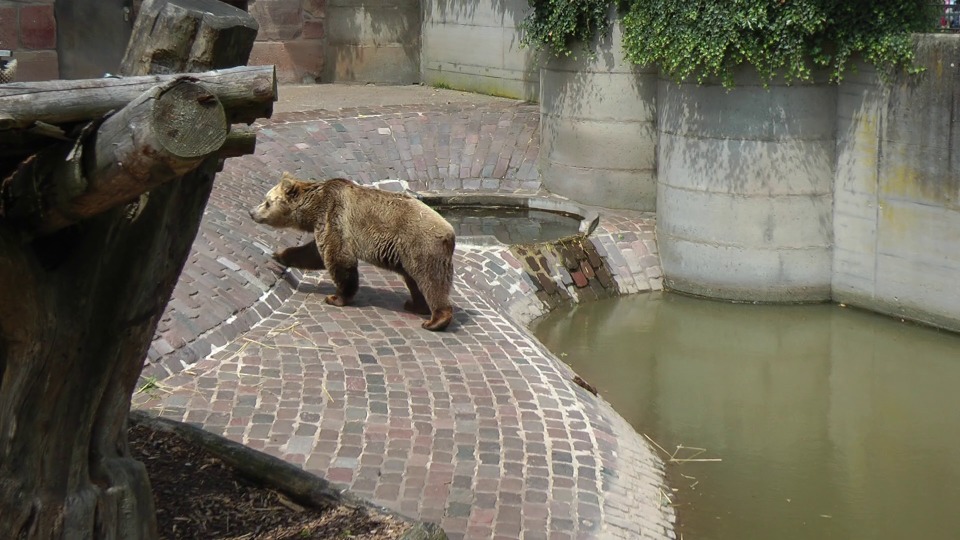}{Image 1} &     
        \labelimage{0.25}{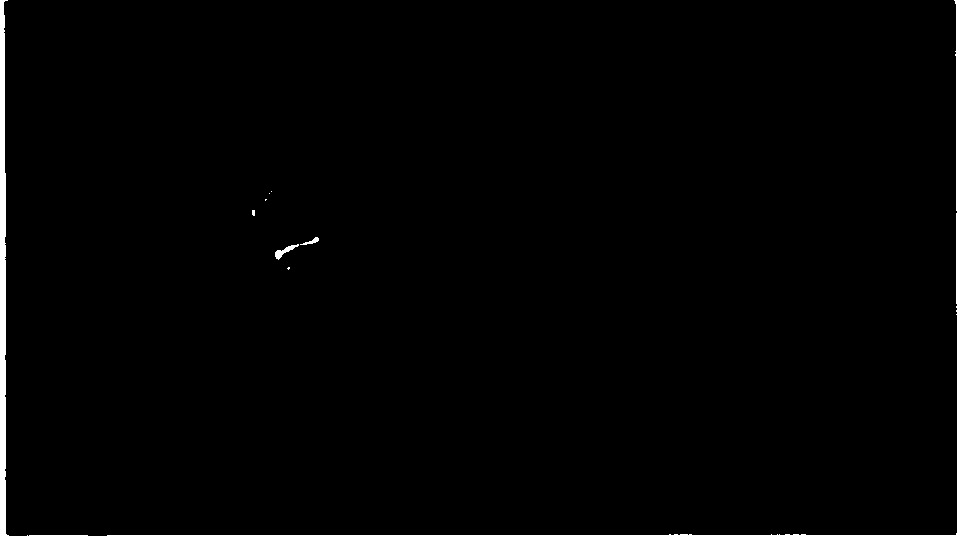}{Occ.} &     
        \labelimage{0.25}{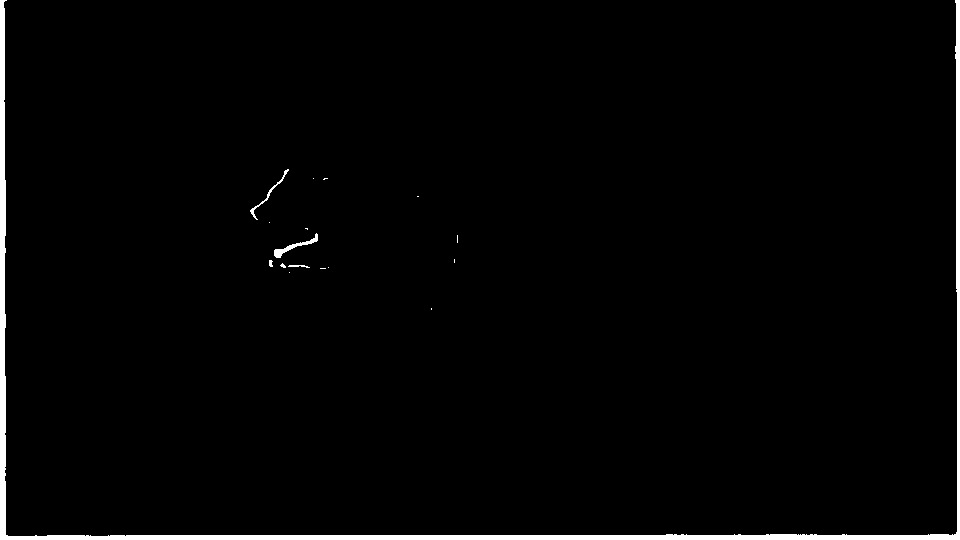}{Occ. (ft-sd)} \\
        \labelimage{0.25}{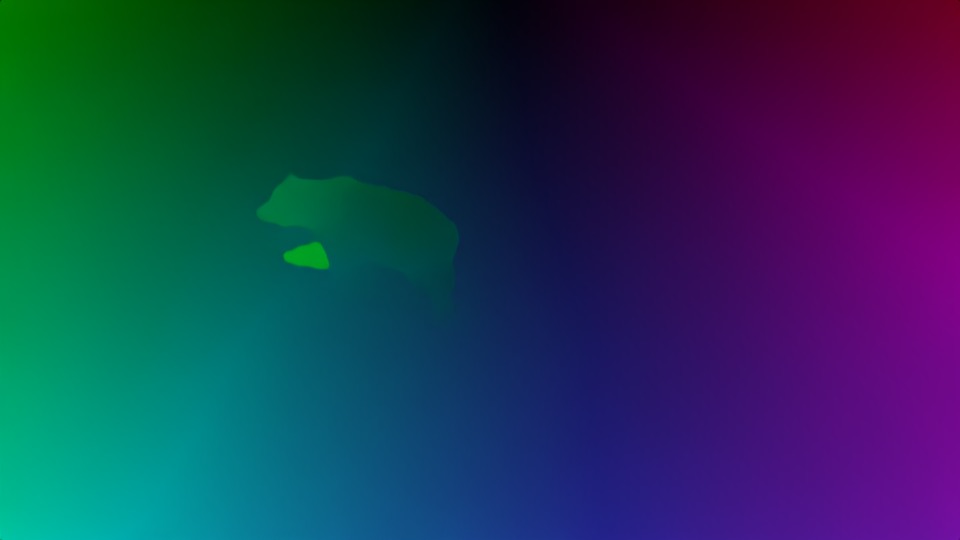}{FlowNet2} &     
        \labelimage{0.25}{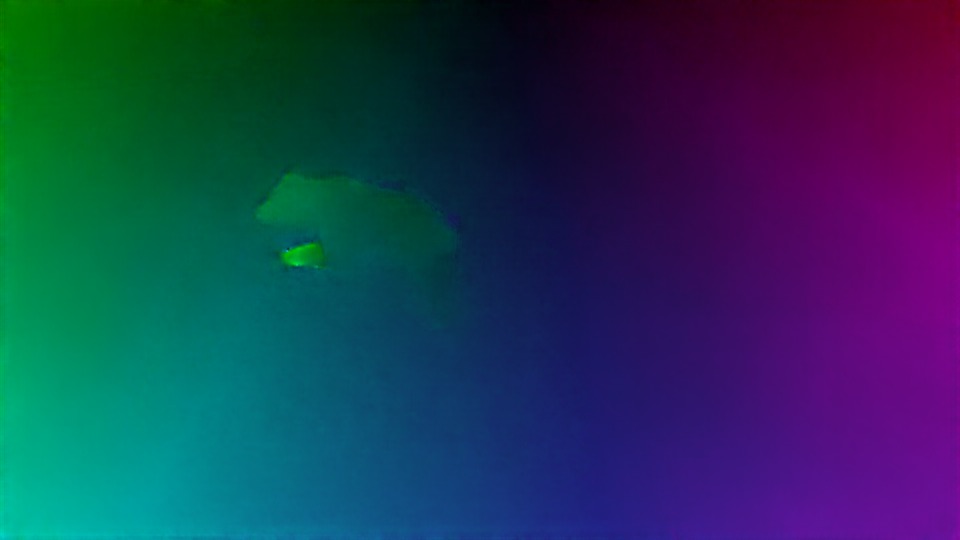}{Flow Ours} &     
        \labelimage{0.25}{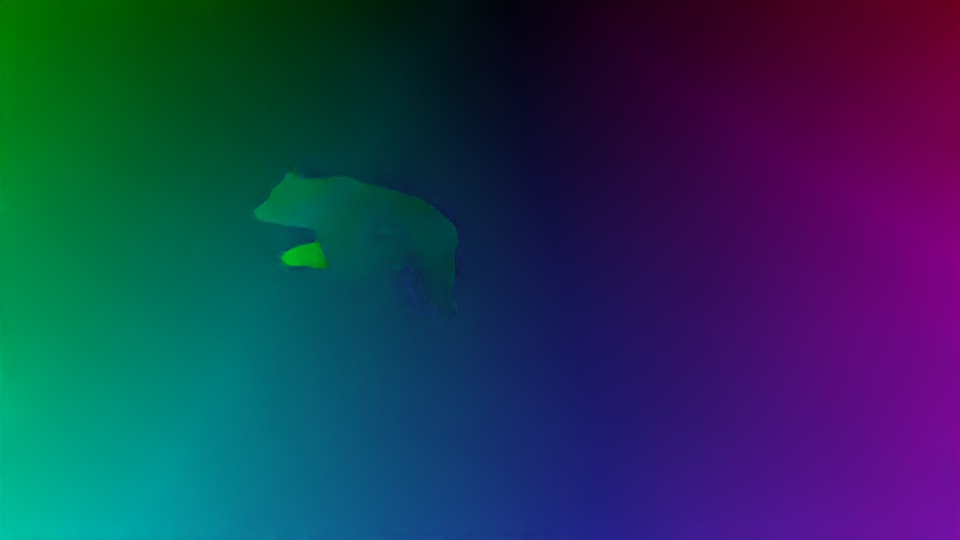}{Flow Ours (ft-sd)} \\
    \end{tabular}  
    \end{center} 
  \caption{
    One more example of our joint motion boundary, occlusion and optical flow estimation
    on some real images. We provide estimations for FlowNet-CSS and FlowNet-CSSR-ft-sd. 
    One can observe that our method provides very sharp estimations and that the 
    version fine-tuned for small displacements provides a bit more details. 
    In this case the motion boundaries estimated from FlowNet-CSSR-ft-sd
    segment the bear from the background very well
    \label{fig:real_gallery3}
  }
\end{figure}

For training CNNs it is common to normalize input and output data to a common range. 
Let $\mathbf{f}$ be the output of the network, $\mathbf{y}^\mathrm{gt}$ the ground-truth and $\mathbf{y}$ our prediction. 
The original FlowNet~\cite{flownet,flownet2} provided the following implementation: 
\begin{equation*}
    \mathrm{min}\,\,\mathcal{L}(\frac{1}{20}\cdot \mathbf{y}^\mathrm{gt}, \mathbf{f}) \mathrm{\,,}
\end{equation*}
\begin{equation*}
    \mathbf{y} = 20\cdot\mathbf{f} \mathrm{\,,}
\end{equation*}
where $\mathcal{L}$ is the loss function. 
In other words, the ground-truth was scaled down by a factor of $20$. This leads to very small values in the network. 
We instead propose to scale up the values inside the network by removing the coefficient:
\begin{equation*}
    \mathrm{min}\,\,\mathcal{L}(\mathbf{y}^\mathrm{gt}, \mathbf{f}) \mathrm{\,,}
\end{equation*}
\begin{equation*}
    \mathbf{y} = \mathbf{f} \mathrm{\,.}
\end{equation*} 
We show a visual example of both scalings in Figure~\ref{fig:fn_samples}. As visible, removing the scaling
produces much better results in the case of small displacements. 
From the first two rows of Table~\ref{tab:training_ablation} we also see that 
changing the scaling has no big effect on the EPE. For disparity, the EPE even decreases a bit. 
Also for the examples from Figures~\ref{fig:real_gallery1},~\ref{fig:real_gallery2} and~\ref{fig:real_gallery3}, our network produces very good results for small displacements even without the small displacement fine-tuning and 
even without the extra small displacement network of FlowNet2~\cite{flownet2}. 

Note that within the decoder of the network the predictions are always upscaled and concatenated 
to further decoder stages. Thus, the optimization needs to scale internal activations to fit the 
ranges of the inserted predictions. 
In general, such a scale can be arbitrary. However, we conjecture that the different 
effects come from weight decay, which can yield different results when operating with different ranges.

\subsection{Weight Maps for Occlusions and Boundaries} 

We predict occlusions by using the cross-entropy loss for the two classes \emph{occluded} and \emph{non-occluded}. We observed that such a network learns to correctly identify large occluded and non-occluded areas, but tends to ignore thin regions. This is because small regions contribute much less to the total loss. 

\begin{figure}[t]
    \begin{center}
    \subfigure[Occlusions]{
        \includegraphics[width=0.4\textwidth]{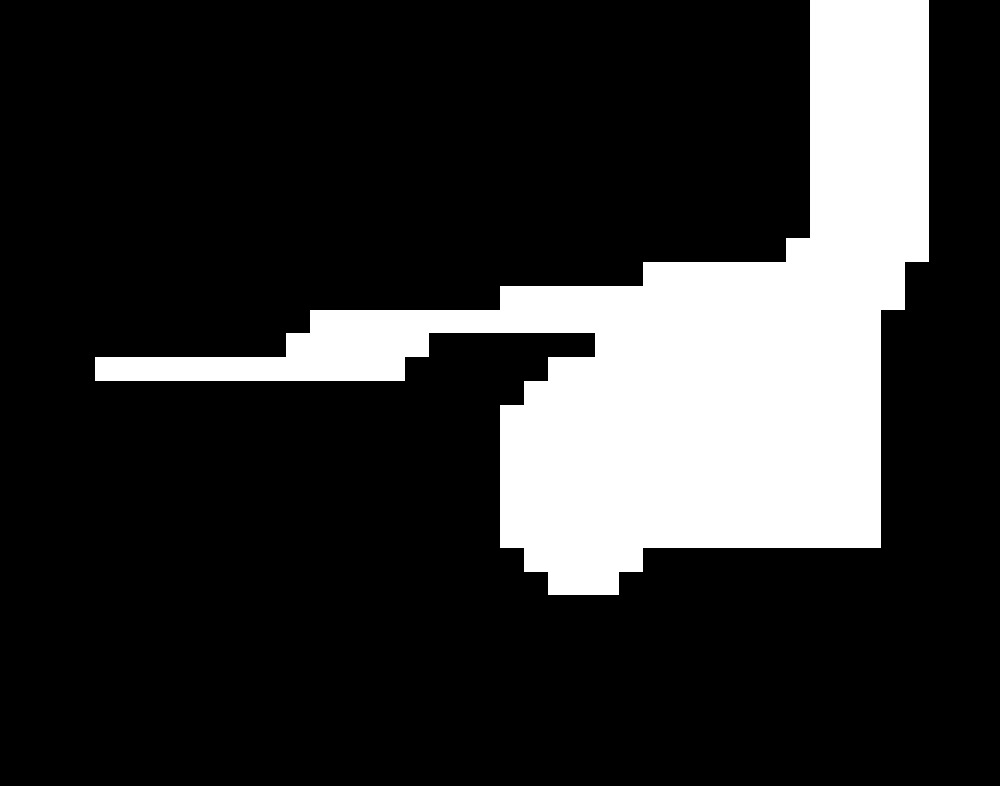}
    }
    \subfigure[Weighting]{
        \includegraphics[width=0.4\textwidth]{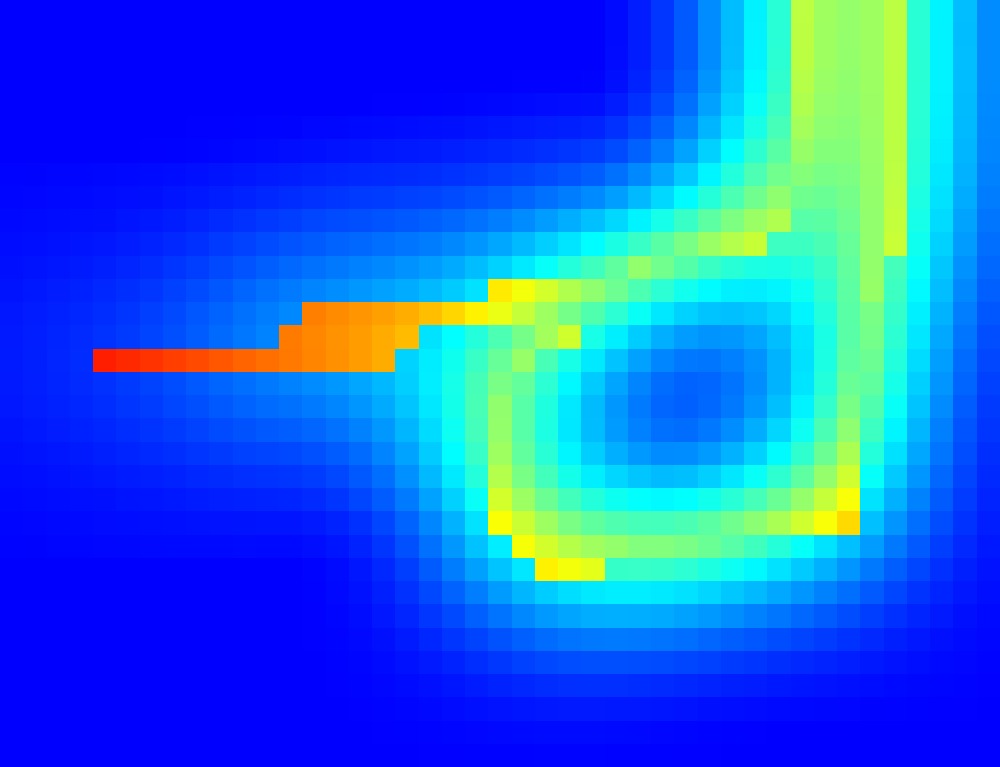}
    }
    \end{center}
    \caption{Illustration of weighting for occlusion and boundary loss.
    Left image shows occlusion ground-truth, right image shows weighting visualized 
    as a heatmap. The weight is higher, the more surrounding pixels have a different occlusion value  
    \label{fig:loss_weighting}
    }
\end{figure}

In general, more non-occluded than occluded pixels are present, thus, the classes are imbalanced. A first solution is to weigh the occluded pixels higher to balance the classes, but this can lead to overflow effects, e.g., pixels of a thin occlusion area have a high weight and are thus more likely to be predicted correctly, but surrounding non-occluded pixels have a low weight and are more easily predicted also as occluded. We propose the following weighting that counteracts this effect:

\begin{equation*} 
    w(x,y)  =  \frac{
        \sum\limits_{i,j \in \mathcal{N}} \delta_{o(x,y) \ne o(i,j)} \cdot g(x - i)g(y - j)
    }
    {
        \sum\limits_{i,j \in \mathcal{N}} g(x - i)g(y - j)
    } \mathrm{\,,} 
\end{equation*}
\begin{equation*}
    \mathrm{with\, } g(d) = e^{-\frac{d^2}{2\sigma^2}} \mathrm{\,,} 
\end{equation*} 
where $\mathcal{N}$ is a neighborhood and $\delta_{o(x,y) = o(i,j)}$ determines whether the neighboring pixel has the same occlusion value as the center pixel. The weight determined for the current pixel is the highest if all surrounding pixels have different occlusion value and decreases as surrounding pixels have similar occlusion value. Neighbors are weighted with a Gaussian with parameter $\sigma$ according to their distance. 
We show an example of the weights 
in Figure~\ref{fig:loss_weighting}. 
For boundary estimation, there is a similar problem. Thus, we apply the same weights. 

\section{Ablation Study of Network Configurations} 

\begin{table}[t]
\vspace*{2mm}
\centering
\begin{tabular}{ |c|c| }\hline 
Correlation level    & EPE       \\ 
\hline 
    $2$       &  $5.62$   \\
    $3$       &  $3.19$   \\ 
\hline
\end{tabular}
\vspace*{4mm}
\caption{Impact of correlation level in DispNetC (the level indicates after which convolution the correlation is defined in the network). We observe a significant performance improvement by just changing the position of the correlation layer to be after the third convolution 
\label{tab:corr}
}
\end{table}

We first change the DispNet architecture by moving the correlation layer up one level. This results mainly in larger strides and correlation distances. From the results of Table~\ref{tab:corr}, one can see that this significantly lowers the endpoint error by almost $50\%$. For an example of improved performance in large displacement regions we refer to the supplementary video. 

In Table~\ref{tab:training_ablation} we provide an ablation study for building our for flow and disparity estimation (in this study we exclude occlusions and boundaries and see only the effects of the stack construction). 

In the first step we show that scaling as proposed in Section~\ref{sec:loss_scaling} has no big effect on the EPE. In the case of disparity it even improves the results. Adding the second network with residual connections~\cite{crl} clearly gives an improvement over the normal stacking. As the second and the third network have similar tasks, it turns out that to train the third network, it is beneficial to copy the weights of the second network for initialization. Although we do not train with the full schedule proposed in~\cite{flownet2}, our final result for FlowNet-CSS is the same. For the case of disparity, our stack significantly improves over DispNetC~\cite{dispnet} (see results in main paper).  

\begin{table}
\centering
\begin{tabular}{ |c|c|c|c|c|c| }
\hline
 Configuration                     & Pred.   & Residual   & Weight & Flow & Disparity \\ 
                                   & scaling & refinement & copy   & EPE  & EPE \\ \hline \hline
C                      & No       & No            & No         & 3.154 & 3.335     \\ \cline{2-2}
C                      & Yes          & No           & No         & 3.208 & 3.194     \\
CS                    & Yes          & No            & No         & 2.340     & 2.634     \\ \cline{3-3}
CS                   & Yes          & Yes          & No        & 2.280    & 2.494     \\ 
CSS                      & Yes          & No            & No         & 2.234     & -     \\ \cline{3-3} 
CSS                & Yes          & Yes          & No        & 2.115    & 2.476    \\ \cline{4-4}
CSS                   & Yes          & Yes          & Yes        & 2.042   &  2.361    \\ 
\hline
\end{tabular}

\vspace*{4mm} 
\caption{Ablation study for training details of a single network and network stacks. Reported errors are from the Sintel train clean dataset. 
We train each network with the schedule of $600k$ iterations on FlyingChairs~\cite{flownet} 
and $250k$ iterations on FlyingThings3D~\cite{dispnet}. For details see text 
\label{tab:training_ablation}
}
\end{table}

\section{Motion Segmentation Results}

In Table~\ref{tab:results_motion_seg} we show results of motion segmentation additionally for the $4$px density evaluation. 
One can observe that the behaviour for the dense version is similar and our approach also performs better in this case. 

\begin{table}[tb]
  \begin{center}
  \begin{tabular}{|l||ccccc|}
    \hline%
    
    Method
    & \multicolumn{5}{c|}{\textbf{Test set} (30 sequences)} \\[0.5mm]
     & {~Density~} & {~Precision~} & {~Recall~} & {~F-Measure~} &  {~\#Objects~} \\
    \hhline{|=#=====|}
    
    {DeepFlow \cite{weinzaepfel:hal-00873592}} & {0.89\%} &88.20\% & {69.39\%} & { 77.67\%}  &  {26/69} \\
    {FlowNet2} & {0.85\%} &{86.73\%} & {68.77\%} & {76.71\%}  & {26/69}\\
    {FlowNet2+occ} & {0.86\%} &{85.67\%} & {70.15\%} & {77.14\%}  & {30/69}\\
    {FlowNetX+occ} & {0.84\%} &{\bf 88.71\%} & {\bf 73.60\%} & {\bf 80.45\%} & {\bf 31/69}\\
    \hline
    {DeepFlow \cite{weinzaepfel:hal-00873592}}  & {3.79\%} &{88.58\%} & {68.46\%} & {77.23\%}  & {27/69} \\
    {FlowNet2} &{3.66\%} &{{87.16\%}} & {68.51\%} & {76.72\%}  & {26/69} \\
    {FlowNet2+occ} &{3.71\%} &{{86.29\%}} & {69.72\%} & {77.13\%}  & {29/69}\\
    {FlowNetX+occ} &{3.61\%} &{\textbf{89.12\%}} & {\bf 72.77\%} & {\bf 80.12\%}  & {\bf 32/69}\\    
    \hline%
  \end{tabular}%
  \end{center}%
 \vspace*{2mm}
 \caption{\label{tab:results_motion_seg} 
  Results of motion segmentation from Keuper~\etal~\cite{keuper15a} on the FBMS-59 test set~\cite{Bro10c,Ochs14} (with 
  sampling densities $8$ and $4$px). 
  The third and seventh rows use flows from FlowNet2~\cite{flownet2} combined with our occlusions. The improved results show that 
  occlusions help the motion segmentation in general. The fourth and last rows show the segmentation using our flow and occlusions, which yields the best performance, also in the $4$px density case 
  }
\end{table}

\section{Training Settings for KITTI}
We fine-tune each network in our stack individually on KITTI. After fine-tuning the first network, we fix the weights and then fine-tune the second network. This process is repeated for all networks in the stack. We train on a mixture of KITTI 2012 and 2015 data, which is split into training and validation sets (with sizes $75\%/25\%$). Each network in the stack is trained for 200k iterations with a base learning rate of $1e-5$. 

The last network in the stack requires motion boundary labels. Since KITTI does not have ground-truth motion boundary labels, we pre-compute the raw motion boundary features estimated from our network
and then tie the predictions to these features using an L2 loss. In this way the network does not forget the already learned boundaries in the case when ground-truth is absent.

\end{document}